\documentclass[lettersize,journal]{IEEEtran}
\usepackage{amsmath,amsfonts}
\usepackage{algorithmic}
\usepackage{algorithm}
\usepackage{array}
\usepackage[caption=false,font=normalsize,labelfont=sf,textfont=sf]{subfig}
\usepackage{textcomp}
\usepackage{makecell}
\usepackage{stfloats}
\usepackage{url}
\usepackage{verbatim}
\usepackage{graphicx}
\usepackage{cite}
\usepackage{multirow}
\usepackage{float}
\usepackage{subcaption}
\usepackage{amsmath,amssymb,amsfonts}
\usepackage{threeparttable}
\usepackage{booktabs}
\usepackage{hyperref}
\usepackage{textcomp}
\usepackage{subcaption}
\usepackage{moresize}
\usepackage{tabularx}  
\usepackage{color}
\usepackage{xcolor}
\usepackage{lipsum}

\begin{document}
\title{Stage-aware Multimodal Brain Graph Contrastive Learning for Neurodegenerative Disorder Diagnosis}


\author{Jingjing~Zhou,~Chenxuan Meng,~Ciyuan Peng,~\IEEEmembership{Member,~IEEE}, Jing Ren,~\IEEEmembership{Graduate Student Member,~IEEE},~and Feng Xia,~\IEEEmembership{Fellow,~IEEE}% <-this % stops a space

\thanks{J. Zhou and C. Meng are with School of Information and Electronic Engineering, Zhejiang Gongshang University, Hangzhou 310018, China (email: zhoujingjing@zjgsu.edu.cn; 22020090019@pop.zjgsu.edu.cn )}% <-this % stops a space
\thanks{C. Peng is with the School of Computer and Network Security, Chengdu University of Technology, Chengdu, 610059, China (e-mail: ciyuan.p@ieee.org)}% <-this % stops a space
\thanks{J. Ren and F. Xia are with School of Computing Technologies, RMIT University, Melbourne, VIC 3000, Australia (e-mail: jing.ren@ieee.org; f.xia@ieee.org)}
\thanks{Corresponding authors: Ciyuan Peng and Jing Ren
}}

%\markboth{IEEE Transactions on Neural Networks and Learning Systems,~Vol.~1, No.~1, January~2015}%

\maketitle

\begin{abstract} 
Neurodegenerative disorders are characterized by their progressive nature, with distinct stages that require precise identification and tracking of progression trajectories for accurate diagnosis. However, existing diagnostic methods often fail to fully capture the unique brain patterns associated with different stages, limiting their ability to effectively target specific stages of the disorders. To address this issue, we propose a Stage-aware Multimodal Brain Graph Contrastive Learning (smBGC) method. Specifically, smBGC introduces a stage-aware feature enhancement strategy to capture stage-specific features across all progression stages, enabling the generation of stage-aware multimodal brain graph representations. In addition, smBGC incorporates a stage-adaptive brain graph augmentation approach within a graph contrastive learning framework, facilitating multimodal fusion and improving the model's capability to distinguish between different disorder stages. Experiments on two real-world datasets demonstrate the superior performance of smBGC in various neurodegenerative disorder diagnosis tasks compared with state-of-the-art methods\footnote{The code is available at:~\url{https://github.com/libertymxc/smBGC}}.
\end{abstract}

% \renewcommand{\abstractname}{Impact Statement}

% \begin{abstract}
% Brain disorders, such as Alzheimer's disease, remain incurable due to the irreversible neuronal damage and pathological changes they cause. However, the disease progression can be slowed down through early-stage detection and pharmaceutical interventions, along with other measures. Unfortunately, most existing methods fail to provide detailed and stage-specific analyses of disease progression, limiting their effectiveness in accurate disease staging and trajectory prediction. Our proposed smBGC framework refines stage-specific representations, enabling a more precise characterization of disease stages. By introducing stage-guided graph enhancement modules within the contrastive learning paradigm, smBGC enhances the model's ability to distinguish between different disease stages. smBGC provides a method capable of distinguishing between stage-specific differences, advancing the field of brain disorders disease diagnosis, and laying the foundation for more precise and personalized disease management strategies. The high performance of smBGC on real-world datasets demonstrates its potential as a valuable tool for both clinical and research applications in the field of brain disorders. Stage-specific analysis of these disorders not only paves the way for more accurate diagnoses but also offers potential for better prediction of disease trajectories, ultimately improving patients' outcomes.
% \end{abstract}


% \fi
\IEEEpeerreviewmaketitle

\section{Introduction}

\IEEEPARstart{N}{eurodegenerative} disorders, such as Alzheimer's disease (AD) and Parkinson's disease (PD), are typical functional impairments of the brain, leading to the decline of various brain functions (e.g., cognitive, emotional, and motor functions)~\cite{chedid2022development}. Notably, neurodegenerative disorders are characterized by the progressive change of neurons and their connections~\cite{CHAKI2023104223}. The hallmark of these disorders is their staged progression, where early stages often present subtle symptoms that gradually worsen over time, ultimately leading to severe functional impairments in later stages~\cite{rasmussen2019alzheimer}. For example, AD patients often experience several transitional stages, including mild cognitive impairment (MCI) and its early and late phases. Therefore, precise stage targeting is an essential aspect of neurodegenerative disorder diagnosis, facilitating early detection, intervention, and tailored treatment.

%stage-aware neurodegenerative disorder diagnosis.

%阶段诊断的重要性

Machine learning-based methods, relying on various data collected by neuroimaging techniques, such as functional magnetic resonance imaging (fMRI), electroencephalography (EEG), and diffusion tensor imaging (DTI), have shown great promise for neurodegenerative disorder diagnosis~\cite{bhandarkar2024deep,10035420,9889239}. Generally, different neuroimaging data provide distinct pathological information in terms of either brain structure or function~\cite{cui2022braingb}. For example, fMRI data records changes in blood oxygenation levels, reflecting brain functional information, while DTI data records the diffusion of water molecules along white matter tracts, reflecting brain structural information~\cite{zimmermann2018unique}. In this context, multimodal learning-based neurodegenerative disorder diagnosis methods have been developed to simultaneously integrate multimodal neuroimaging data (e.g., fMRI and DTI), offering comprehensive structural and functional information to enhance diagnostic accuracy~\cite{bazarbekov2024review}. More recently, due to the capability of effectively modeling complex brain connectivity, multimodal graph learning has emerged as a promising approach for neurodegenerative disorder diagnosis~\cite{dai2019disrupted,11334033,10466590}. Many studies utilize multimodal brain graphs, including structural brain graphs (SBGs) and functional brain graphs (FBGs) constructed from structural imaging data (e.g., DTI) and functional imaging data (e.g., fMRI), respectively, to model complex interactions within the brain and improve model performance~\cite{zhu2022joint,yang2023mapping,ye2024rh}. 
%GNNs运用于脑疾病诊断的背景

% Although current multimodal brain graph learning methods show great potential in neurodegenerative disorder diagnosis, several challenges remain to be addressed. Firstly, most existing methods fail to explore the distinct brain patterns associated with different stages of neurodegenerative disorders~\cite{cho2024neurodegenerative}. Specifically, current methods are often limited to distinguishing between only two stages (e.g., AD and normal stages), failing to comprehensively capture stage-specific features across the entire progression spectrum~\cite{li2021braingnn,song2022multicenter,zhang2022classification}. This inability to accurately identify a patient's specific stage significantly hinders early medical screening and diagnosis. In addition, nodes (brain regions) in multimodal brain graphs usually exhibit extremely complex and high-dimensional features, which often include noise irrelevant to diagnosis~\cite{peng2024adaptive}. This can lead to overfitting and significantly degrade the model's performance.

Although current multimodal brain graph learning methods demonstrate promising potential in the diagnosis of neurodegenerative disorders, several critical challenges remain unresolved~\cite{RenGL4AA2023}. One major limitation lies in their inadequate ability to distinguish among different stages of disease progression~\cite{cho2024neurodegenerative}. Most existing approaches primarily focus on binary classification (e.g., AD vs. normal controls), neglecting the nuanced and stage-specific alterations that occur across the full continuum of neurodegeneration~\cite{li2021braingnn,song2022multicenter,zhang2022classification}. This lack of fine-grained stage differentiation not only restricts our understanding of disease mechanisms but also severely limits the effectiveness of early screening and intervention. Early-stage diagnosis is particularly crucial, as therapeutic strategies are often more effective before irreversible neurological damage occurs. Without accurate identification of early or intermediate stages, opportunities for timely medical intervention may be missed. Moreover, the complex, high-dimensional nature of features in multimodal brain graphs—often containing substantial noise unrelated to disease—further challenges model robustness and increases the risk of overfitting~\cite{peng2024adaptive}, ultimately compromising diagnostic accuracy.

%多模态脑疾病诊断方法的背景和缺陷

To address these issues, this paper proposes a novel Stage-aware Multimodal Brain Graph Contrastive Learning method, namely smBGC. smBGC effectively distinguishes different stages of neurodegenerative disorders by exploring distinct brain patterns within each stage. We first construct multimodal brain graphs, containing SBGs and FBGs derived from DTI and fMRI data, respectively. Then, smBGC introduces a stage-aware feature enhancement strategy to amplify stage-specific features and obtain stage-aware multimodal brain graphs, as well as reduce noise. Specifically, this strategy consists of two key steps: communal feature dropping (CFD) and stage feature extraction (SFE). In CFD, common features across all stages are treated as noise and dropped to avoid interference. In SFE, stage-specific features are extracted for each stage to obtain stage-aware SBGs and stage-aware FBGs, which are treated as two views of brain graphs. Following this, we propose a stage-adaptive brain graph augmentation approach, which strengthens stage-specific attributes by applying a node masking technique based on stage features. Afterwards, we employ contrastive learning to facilitate multimodal fusion, jointly embedding structural and functional features.

This work provides insights into the progression-related characteristics of neurodegenerative disorders and investigates specific brain patterns associated with different disorder stages. Notably, our method demonstrates strong capabilities in identifying early-stage abnormalities, which is critical for enabling timely intervention before significant neural deterioration occurs. The main contributions of this paper are as follows:
\begin{itemize}
    \item[$\bullet$] We propose a Stage-aware Multimodal Brain Graph Contrastive Learning (smBGC) method to uncover brain patterns across different stages of neurodegenerative disorders. smBGC captures stage-specific patterns and provides discriminative representations for different stages of neurodegenerative disorders.
    \item[$\bullet$] We design a stage-aware feature enhancement strategy, which comprises a communal feature dropping module and a stage feature extraction module. This strategy amplifies brain graph features on each stage, enabling the construction of stage-aware multimodal brain graphs.
    \item[$\bullet$] Extensive experiments on two real-world datasets demonstrate the proposed model outperforms state-of-the-art methods, highlighting its superiority in various neurodegenerative disorder diagnosis tasks. %\textcolor{red}{To facilitate reproducibility and further research, we have made our implementation and datasets publicly available at: \url{https://github.com/libertymxc/smBGC}.}
\end{itemize}

The rest of the paper is organized as follows. Section~\ref{related_work} reviews related works. Section~\ref{preliminaries} gives the preliminaries. Section~\ref{method} details the proposed smBGC. Section~\ref{experiment} presents the experimental setting and results. Finally, we conclude our work in Section~\ref{conclusion}.

\section{Related Work}\label{related_work}

\subsection{Neurodegenerative Disorder Diagnosis}

Most current machine learning-based neurodegenerative disorder diagnosis methods primarily focus on binary classification tasks to distinguish between two different stages. For instance, for Alzheimer’s
disease, a common task is classifying between two stages, such as normal control (NC) vs. MCI, NC vs. AD, and MCI vs. AD~\cite{yang2023mapping,bian2023adversarially}. Song \textit{et al.} \cite{song2022multicenter} integrated fMRI and DTI data to construct brain graphs while also integrating non-imaging information for early AD diagnosis, showing good performance in six binary classification tasks. Ma \textit{et al.} \cite{10294288} focused on distinguishing between stages of MCI (e.g., early MCI vs. MCI and MCI vs. late MCI), proposing a novel multi-graph cross-attention region-aware feature fusion network with various brain region templates. Yang \textit{et al.} \cite{yang2023mapping} used cross-modal mutual learning to complete binary classification tasks like NC vs. MCI, MCI vs. AD, and NC vs. AD in AD detection, and NC vs. PD in PD detection. Cai \textit{et al.} \cite{10214974} proposed a network manifold harmonic discriminant analysis (MHDA) method for accurately detecting AD binary stages.

Stage targeting of neurodegenerative disorders, such as AD and PD, is crucial for early diagnosis and treatment planning. Traditional binary classification between only two stages is insufficient. Additionally, differences in brain state across different stages of patients are often neglected, which may lead to performance drops in methods. Recently, a few studies have focused on the diagnosis of more stages~\cite{peng2024stage}. Cho \textit{et al.} \cite{cho2024neurodegenerative} introduced a node-variant convolution mechanism that adaptively captures both localized homophily and heterophily to learn the progressive degenerative changes in brain graphs, enabling the distinction of five AD stages and three PD stages. Hu \textit{et al.} \cite{hu2024cost} developed a contrastive learning framework based on functional connectivity (FC) and high-order functional connectivity (HOFC) to address class imbalance in diagnosing five AD stages. However, these approaches primarily rely on single-modality data, indicating a need for further research on precise diagnosis and the learning of features across different modalities.

\subsection{Brain Graph Analysis}

Graph learning-based brain graph analysis methods focus on two main types of graphs: population graphs and brain graphs~\cite{10492652,zhang2022classification,yu2025long}. Population graphs are constructed based on population, where each node represents an individual’s brain. These graphs learn relationships between subjects effectively, modeling non-imaging information like gender and age as weighted edges. Population-GCN \cite{parisot2017spectral} represents a set of participants as a sparse graph, with nodes associated with image-based feature vectors, and edges representing phenotype information and feature similarity among participants. To address the limitation of ignoring biological biomarkers, Hao \textit{et al.} \cite{zhang2022classification} proposed a local-to-global graph neural network (LG-GNN), which learns feature embeddings for local brain regions and identifies biomarkers, followed by a global Subject-graph neural network that learns relationships between embeddings generated by the local ROI-GNN and non-imaging information. Song \textit{et al.} \cite{song2022multicenter} integrated multimodal brain imaging data to construct nodes, designing a multi-center attention graph to account for non-image information and feed the graph into a proposed multi-channel mechanism to enhance model performance. %Cho \textit{et al.} \cite{cho2024neurodegenerative} proposed a cost-sensitive weighted contrastive learning method based on GCN (CSWCL-GCN) to address class imbalance and heterogeneity in multi-class AD classification. Using a dual-channel GCN and multi-view population graphs, the method integrates class prototype-based weighting for negative pairs and a cost-sensitive mechanism to balance class contributions during training.

Methods using brain graphs treat each region of interest (ROI) in the brain as a node and the connectivity between ROIs as edges. Ma \textit{et al.} \cite{10233901} proposed a multi-scale dynamic graph learning (MDGL) framework that leverages complementary information across spatial resolutions and temporal dynamics of rs-fMRI data to enhance the detection of brain disorders. Liu \textit{et al.} \cite{10155767} embedded the guiding priors from multiscale brain atlases into the hierarchical graph convolutional neural networks for brain disorder diagnoses. Some studies focus on graph contrastive learning for brain graph analysis. Xu \textit{et al.} \cite{10508252} proposed ContrastPool, a contrastive graph pooling method with dual attention mechanisms tailored for brain networks and fMRI characteristics. HSGRL \cite{9950427} is a symbolic graph contrastive learning model with an interpretable graph pooling module that constructs subgraphs by computing nodes with high information gain. Peng \textit{et al.} \cite{peng2024adaptive} proposed an adaptive brain network enhancement method based on group-aware graph learning to construct group-aware brain graphs, which can adapt to distinct groups, reducing the interference of noise. HGFM \cite{10967508} presents a self-supervised pretraining framework for brain disease diagnosis that models high-order correlations in brain networks using a hypergraph paradigm. DHANet \cite{11010081} employs a 3D dynamic ConvMixer for feature extraction, dynamic prototype learning for region identification, and joint dynamic edge-correlation modeling to construct a heterogeneous-aware structural brain network.

Different from the above-mentioned papers, this paper treats two modalities (structure and function) from the same patient as different perspectives of the brain graph, designing enhancement strategies using connection profiles as node features for the brain graph and employing a contrastive learning framework for multimodal fusion.

\subsection{Graph Contrastive Learning}

Graph contrastive learning (GCL) methods create multiple graph views through random augmentations, optimizing representations by contrasting positive and negative samples \cite{RenEAGLE2023}. The aim is to minimize the distance between similar objects (positive pairs) in the embedding space and maximize the distance between dissimilar objects (negative pairs). 
GCL methods are classified into intra-scale contrast, focusing on comparisons within the same level, and inter-scale contrast, examining comparisons across different levels. Intra-scale contrast focuses on comparisons at the same level, such as between node-level views (local-local). GCA \cite{zhu2021graph} refines graph augmentation methods, incorporating topological and semantic prior information to address the performance limitations of single augmentation schemes. GPA \cite{10499878} explores the impact of different augmentation strategies on graph instances by iteratively updating the personalized augmentation selector, and assigning personalized augmentation types based on the characteristics of each graph during training. ASP \cite{chen2023attribute} diverges from traditional two-view comparisons by considering three different graph views—original, attribute, and global structure—and performs contrastive learning across these views in a unified manner. DCLN \cite{10097557} presents a dual curriculum contrastive module to enhance graph structural information and reduce feature redundancy with two target functions, preventing collapse at the node and feature levels. 

Inter-scale contrast, on the contrary, focuses on global-to-local comparisons between node-level views and graph-level views. For example, InfoGraph \cite{Sun2020InfoGraph} contrasts representations at graph-level and sub-structural scales (nodes, edges, triangles, etc.). Zhao \textit{et al.} \cite{10505804} proposed an augmentation-free graph contrastive transformer optimized by the nearest neighbors to avoid model degradation caused by augmentation. In this work, we use a global-to-local contrast mode to differentiate between structural views and functional views, contrasting node representations from one view with graph representations from another view, with mutual information calculation using the Jensen-Shannon Divergence (JSD)~\cite{DBLP:conf/iclr/HjelmFLGBTB19}. %DGI \cite{veličković2018deep} is a method for node-level embedding and graph-level representation contrast by maximizing the mutual information between local representations and their corresponding global summaries to train the model.

% Recently, some works have explored multimodal contrast scenarios. InfoGCL \cite{xu2021infogcl} presents an information-aware graph contrastive learning framework that reduces mutual information between contrastive parts while maintaining task-relevant information, thus minimizing information loss in the graph representation learning process. MS-CIA \cite{zhang2023multi} optimizes objectives at three different scales, including graph, subgraph, and node, to learn node representations. TopoGCL \cite{chen2024topogcl} introduces topological invariance and extended persistence to GCL, proposing a new contrastive mode that focuses on topological representations at multiple resolutions. In this work, we use a global-to-local contrast mode to differentiate between structural views and functional views, contrasting node representations from one view with graph representations from another view, with mutual information calculation using the Jensen-Shannon Divergence (JSD)~\cite{hjelm2019learningdeeprepresentationsmutual}.

Recent advances in GCL have introduced several innovative frameworks. Wang \textit{et al.} \cite{10.24963/ijcai.2023/260} proposed iDCL, a novel GCL framework that disentangles user intents and behavior distributions in user-item-concept graphs, significantly improving recommendation performance and interpretability. Jiang \textit{et al.} \cite{10.1145/3580305.3599768} developed AdaGCL, which employs trainable generative and denoising view generators to enable adaptive, data-driven augmentation, thus improving robustness against noise and sparsity. Zhuo \textit{et al.} \cite{10143747} introduced an innovative contrastive learning approach that strengthens GNNs by injecting long-range similarity information through dynamically updated augmented views, effectively overcoming the limitations of static augmentations in unsupervised graph representation learning. Further advancing the field, Li \textit{et al.} \cite{10058898} proposed iGCL, an augmentation-free framework that eliminates the need for negative samples while ensuring representation invariance and discriminability through a novel invariant-discriminative loss function.

\begin{table}[H]
	\centering
	\caption{Notations in this paper.}
    \setlength{\tabcolsep}{6pt} % 可根据需要调整列间距
    \renewcommand{\arraystretch}{1.2} % 增加行距让表格更美观
\begin{tabular}{p{1.5cm}|p{7cm}}
\toprule
\textbf{Notations} & \textbf{Description}  \\ \midrule
$B$  & Dataset of brain graphs \\
$G_i^s$, $G_i^f$, $G_i^*$ & SBG, FBG and a brain graph from any modality of the i-th subject, respectively \\
$y^i$ & The stage label of i-th subject \\
$\mathcal{V}$ & Node set of i-th subject \\
$E_{i}^{*}$ & Edge set of $G_i^*$, $*$ represents any modality \\
$M$ & The number of subjects in $B$ \\
$N$ & The number of nodes \\
$D$, $D^{\prime}$ & The number of node features' dimensions and the number of dimensions after denoising, respectively \\
$\lambda$ & The number of dropped features on each side (top and bottom), the total number of dropped features is 2$lambda$ \\
$K$ & The number of groups or stages \\ \midrule
$\mathbf{A}_{i}^{*}$ & Weighted adjacency matrix of $G_i^*$ \\
$\mathbf{X}_i^*$ & Node feature matrix of $G_i^*$, $\mathbf{X}_i^* = [\mathbf{x}_1^i,\mathbf{x}_2^i,...,\mathbf{x}_N^i]^\top$ \\
$\mathbb{X}$ & The node feature set of a group of subjects, $\mathbb{X} = \{\mathbf{X}_1,\mathbf{X}_2,...,\mathbf{X}_M\}$ \\
$\widetilde{\mathbb{X}}, \hat{\mathbb{X}}$ & The reshaped node feature set and the denoised node feature set, respectively, $\widetilde{\mathbb{X}} = \{\widetilde{\mathbf{X}}_1,...,\widetilde{\mathbf{X}}_N\}$ and $\hat{\mathbb{X}} = \{\mathbf{\hat{X}}_1,...,\mathbf{\hat{X}}_M\}$ \\
$\widetilde{\mathbf{X}}, \mathbf{\hat{X}}$ & The matrix containing node features across all subjects and the denoised node feature, respectively \\
$\mathbf{S}$ & The feature scoring matrix \\
$\mathbf{S}_c$ & The common scoring matrix\\
$\mathbf{F}_{k}$ & The stage feature scoring matrix of stage $k$ \\
$\mathbf{F}_{k}^{f}$, $\mathbf{F}_{k}^{s}$ & The functional and structural stage feature scoring matrix of stage $k$ \\
$\mathbb{F}$ & The collection of stage feature scoring matrices, $\mathbb{F} = \{\mathbf{F}_{k}\}_{k=1}^{K}$ \\
$W$ & Weighted vector \\
$\hat{\mathbf{X}}^{*\prime}$ & The augmented node feature matrix \\
\bottomrule
\end{tabular}
\label{tab:notations}
\end{table}

\section{Preliminaries}\label{preliminaries}

This paper primarily addresses the problem of stage-aware multimodal brain graph analysis. Given a brain graph dataset $B=\{(\{G_i^s,G_i^f\},y^i)\}_{i=1}^M$, where $G_i^s$ and $G_i^f$ represent the SBG and FBG of the $i$-th subject, respectively, $y^i$ is the corresponding disorder label, and $M$ is the number of subjects. Each modality's brain graph is constructed based on region of interest (ROI) connectivity, which can be described by a weighted graph $G_{i}^{*}=(\mathcal{V},E_{i}^{*},\mathbf{A}_{i}^{*})$, where $\mathcal{V}=\{v_{j}\}_{j=1}^{N}$ represents a set of $N$ nodes (ROIs), $E_{i}^{*}=\mathcal{V}\times\mathcal{V}$ is the edge set, $\mathbf{A}_{i}^{*}\in\mathbb{R}^{N\times N}$ represents the weighted adjacency matrix. Notably, the node set $\mathcal{V}$ remains the same across different subjects. The node feature matrix of $G_i^*$ is $\mathbf{X}_i^*=$$[\mathbf{x}_1^i,\mathbf{x}_2^i,...,\mathbf{x}_N^i]^\top$, where $\mathbf{x}_j^i\in\mathbb{R}^D$ represents the feature vector of node $v_j$ in $G_i^*$. Specifically, we directly adopt the weighted adjacency matrix as the node feature matrix, i.e., $X_i^* = A_i^*$, such that the feature vector $\mathbf{x}_j^i$ of node $v_j$ corresponds to its connectivity profile (the $j$-th row of $A_i^*$), and $D=N$. For all brain graphs in the dataset, $\{\{G_1^s,G_1^f\},\{G_2^s,G_2^f\},...,\{G_M^s,G_M^f\}\}$, they can be grouped into $K$ groups, each representing a stage of the neurodegenerative disorder. The main goal of this paper is to utilize multimodal brain graphs as multiple views for contrastive learning, training a GNN encoder  $g(\cdot)$, to learn a low-dimensional latent representation for each subject, thereby obtaining a graph representation that integrates functional and structural information of brain for disorder diagnosis and stage targeting.

Formally, we calculate noise features using all brain graphs, dropping them to obtain a processed node feature matrix $\mathbf{X}_i^{*\prime}$, then for the $k$-th stage, we extract stage features $\mathbf{F}_k$ from all group members $\left\{\left\{G_1^s,G_1^f\right\}_k,\left\{G_2^s,G_2^f\right\}_k,...,\left\{G_j^s,G_j^f\right\}_k\right\}$, where $j$ is the number of subjects belonging to stage $k$. Subsequently, stage features are used to enhance the model's learning, embedding the brain graphs into a specific stage space. All the notations in this paper are listed in Table \ref{tab:notations}.

\section{Methodology}\label{method}

% In this section, we present our smBGC approach with the overall architecture illustrated in Fig. \ref{overview}. We first explain how smBGC employs the feature enhancement strategy to obtain stage-aware multimodal brain graphs. Next, we describe the process of our stage-adaptive brain graph augmentation approach. Finally, we detail the multimodal fusion via contrastive learning.

\begin{figure*}[!t]
	\centering
	\includegraphics[width=\textwidth]{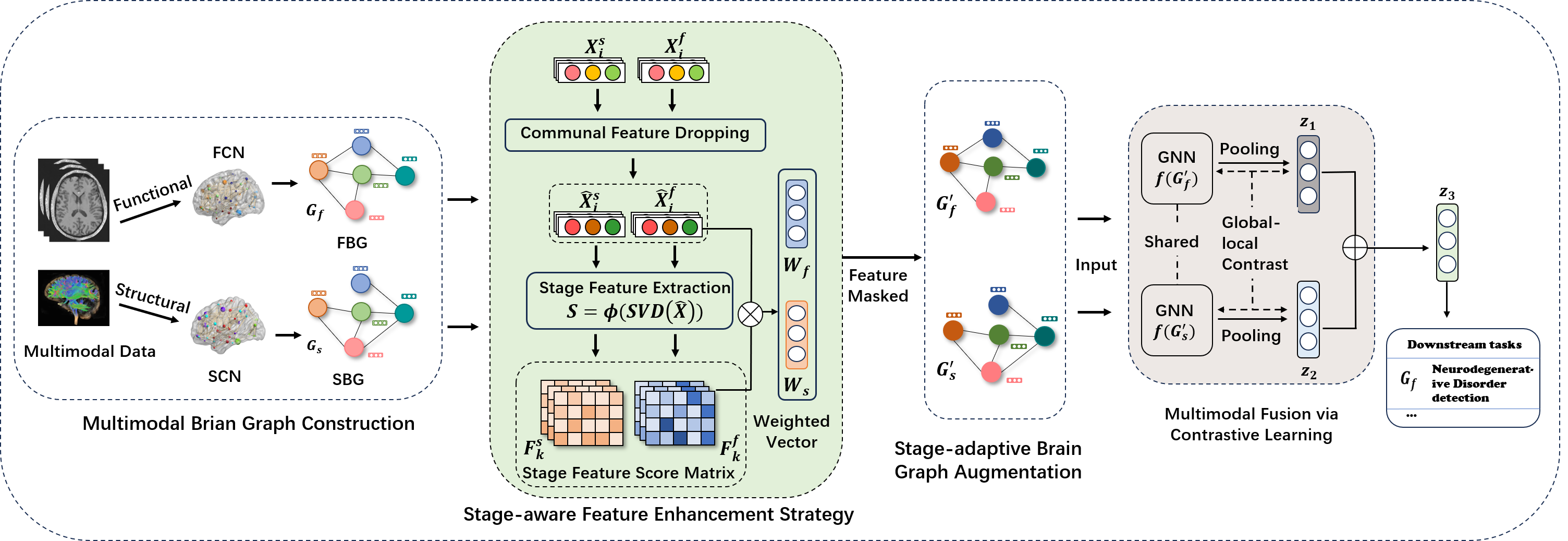}
	\caption{The overall framework of smBGC. Multimodal data are used to construct the FBG and SBG. The feature score matrix of each stage is generated using SVD, and feature enhancement is performed with weighted vectors. Both structural and functional brain graphs undergo feature masking and enhancement, with the new graph representations input into a parameter-sharing GNN for contrastive learning to capture the associations and differences between FBGs and SBGs.}
	\label{overview}
\end{figure*}

In this section, we present our smBGC approach, with the overall architecture illustrated in Fig. \ref{overview}. We first explain how smBGC employs a stage-aware feature enhancement strategy to adaptively construct brain networks for patients at different stages, thereby reducing noise features while enhancing stage-specific characteristics. Next, we describe our stage-adaptive brain graph augmentation approach, which refines the multimodal network representations. Finally, we detail the multimodal fusion through contrastive learning, in which structural and functional brain networks are treated as multiple perspectives of the brain and jointly embedded into a unified space. 
% The training and testing procedures are provided in Algorithm~\ref{alg:train} and~\ref{alg:test}.

 \begin{figure}[h]
	\centering
	\includegraphics[width=\columnwidth]{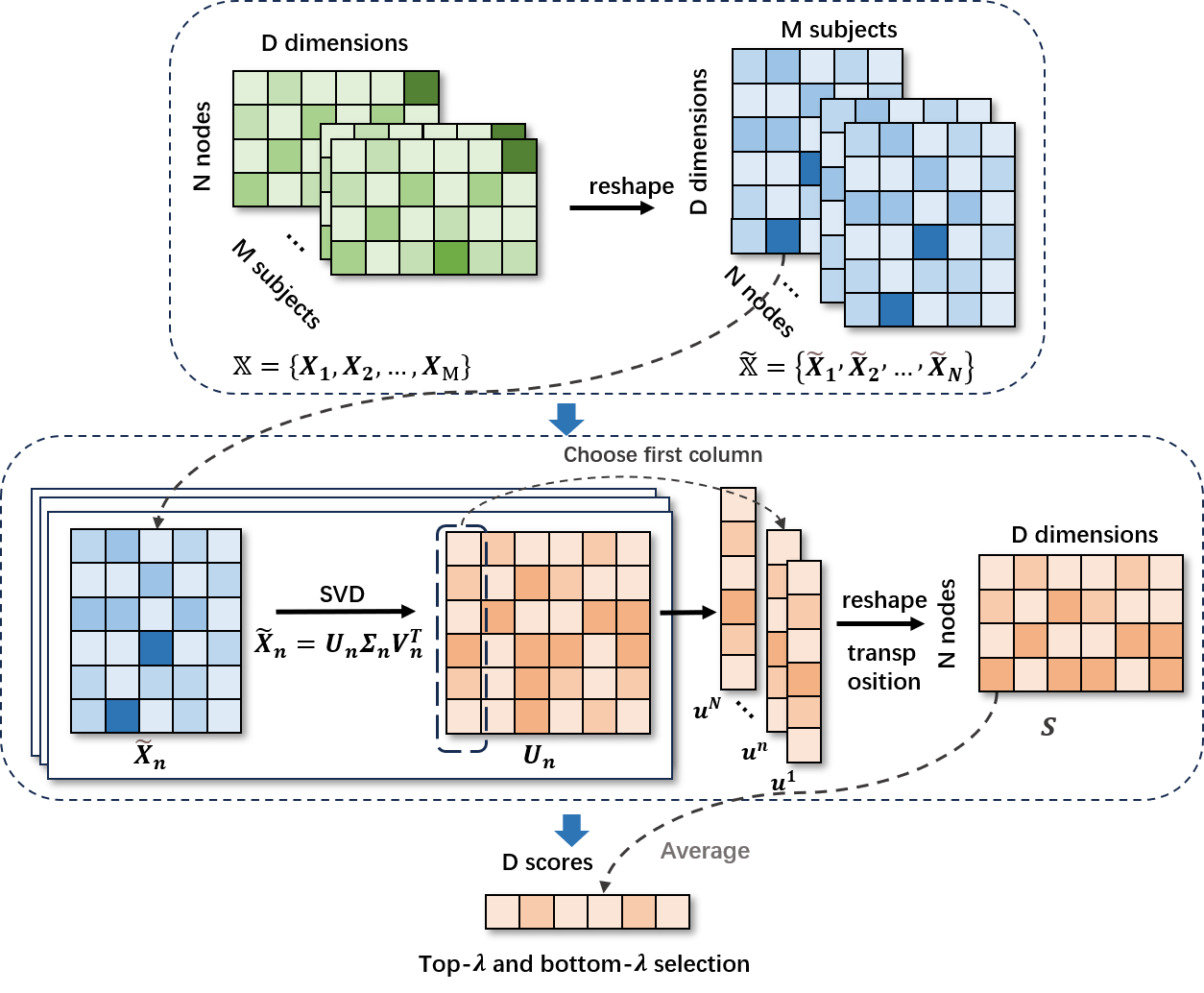}
	\caption{Overview of the proposed Communal Feature Dropping.}
	\label{common}
\end{figure}

\subsection{Stage-aware Feature Enhancement Strategy}
% In this section, we detail how to obtain stage-aware brain graphs, treating the connectivity profile of patients as node features of the brain graph, which undergoes two processing steps: communal feature dropping (CFD) and stage feature extraction (SFE).

In this section, we detail how to obtain stage-aware brain graphs using the stage-aware feature enhancement strategy. The key to this strategy is a Singular Value Decomposition (SVD)-based feature scoring function, which computes the importance score of each feature from a cohort of patients' brain graphs, thereby reflecting the node feature's significance within that subgroup. This scoring framework enables two critical processing stages: Communal Feature Dropping (CFD) to eliminate non-discriminative features across stages, and Stage Feature Extraction (SFE) to selectively capture discriminative features associated with specific disorder stages.

\subsubsection{SVD-based Node Feature Scoring Function}
The scoring function is inspired by the Latent Semantic Indexing (LSI) algorithm in natural language processing, which fundamentally employs SVD on a term-document matrix (with elements representing word frequencies or weighted values) to extract latent semantic structures: the left singular matrix characterizes the distribution of terms in the latent space, reflecting their semantic properties (such as importance or topic relevance), while the right singular matrix corresponds to the semantic patterns of documents.
In this study, we use patients' connectivity matrices as node features and reorganize the node feature matrices of a group of subjects to obtain a feature-node matrix analogous to the term-document matrix. Given their similar high-dimensional characteristics and entity-relationship structures, we can adapt this core concept of LSI to brain network analysis. By decomposing this matrix, we achieve dimensionality reduction and employ the left singular vectors to interpret key nodal features of the group, mirroring the approach of understanding term importance through their weight distribution in semantic space.

Specifically, taking an arbitrary group of subjects as an example, to introduce the subject more concisely, we consider only the brain graphs from a single modality, omitting the superscripts that denote specific modality graphs. Starting from the node features of this group $\mathbb{X}=\{\mathbf{X}_1,\mathbf{X}_2,...,\mathbf{X}_M\}$, where $\mathbf{X}_\mathrm{m}\in\mathbb{R}^{N\times D}$ represents the node feature matrix of the $m$-th subject. The first step of the scoring function is to rearrange the features of each node across all $M$ subjects of this group, resulting in a new set of feature-node matrices $\widetilde{\mathbb{X}}=\{\widetilde{\mathbf{X}}_1,\widetilde{\mathbf{X}}_2,...,\widetilde{\mathbf{X}}_N\}$, where $\widetilde{\mathbf{X}}_n\in\mathbb{R}^{D\times M}$ denotes the matrix containing node features across all subjects of node $n$. We then perform SVD on each of these new matrices to score the features at each node. The first column of the left singular matrix obtained from SVD serves as the score for each feature, representing the importance of each feature to each node within this group. Thus, we obtain a feature score matrix for all nodes, which can be expressed as follows:
\begin{eqnarray}
\widetilde{\mathbf{X}}_{n}=\mathbf{U}_{n}\mathbf{\Sigma}_{n}\mathbf{V}_{n}^{T},\quad n=1,2,...,N,
\end{eqnarray}
\begin{eqnarray}
\mathbf{u}^n=\mathbf{U}_n[:,1]=\begin{bmatrix}\boldsymbol{\mu}_1^n,\boldsymbol{\mu}_2^n,...,\boldsymbol{\mu}_D^n\end{bmatrix}^T,
\end{eqnarray}
\begin{eqnarray}
\mathbf{S}=[\mathbf{u}^1,\mathbf{u}^2,...,\mathbf{u}^N]^T=\phi(\mathbb{X}).
\end{eqnarray}
Here, $\mathbf{U}_n$ is the left singular matrix, $\mathbf{u}^n$ is the set of feature scores for a single node, $\boldsymbol{\mu}_1^d$ is the score of the $d$-th dimensional feature, and  $\mathbf{S}\in\mathbb{R}^{N\times D}$ is the communal feature score matrix for all nodes in the group. For better understanding, we provide the process of the SVD-based scoring function (as shown in Algorithm~\ref{algo1}).

Note that prior to constructing the feature-node matrix $\widetilde{\mathbf{X}}_n$, we take the absolute value of functional connectivity values (derived from Pearson correlation and thus signed), to ensure that the scoring reflects connection strength rather than confounded by correlation sign. Structural connectivity values, non-negative by construction, require no such transformation. The resulting score vector is taken directly from the left singular vector without further normalization.

For a single matrix of size $p \times q$, the time complexity of performing Singular Value Decomposition (SVD) is $O(\min(p \times q^2,\ p^2 \times q))$ and the space complexity is $O(p \times q)$. In Algorithm~\ref{algo1}, SVD is applied to each node-level matrix.

$\widetilde{\mathbf{X}}_n\in\mathbb{R}^{D\times M}$. Accordingly, the computational complexity for each node is $O(\min(D\times M^2,D^2\times M))$, and the space complexity is 
$O(D\times M)$. Considering that there are $N$ brain regions (nodes) in total, the overall time and space complexities are:
\begin{eqnarray}
T=O(N\times \min(D\times M^2,D^2\times M)),
\end{eqnarray} 
\begin{eqnarray}
S=O(M\times N\times D).
\end{eqnarray} 
Given that the number of nodes $N$ is relatively small in standard brain atlases (e.g., AAL90), and that neurodegenerative disorder datasets are typically of limited scale, this computational cost is acceptable in practice. In addition, SVD operations can be efficiently parallelized on modern GPUs or multi-core CPUs.
When extending the framework to higher-resolution brain atlases (e.g., AAL116 or CC200) or longitudinal multi-session datasets, the computational burden can be further reduced using truncated SVD or randomized SVD, which provide efficient low-rank approximations while maintaining interpretability of the feature representations.

\subsubsection{Communal Feature Dropping}

We achieve brain network denoising by identifying and processing communal features. Specifically, we treat all subjects across different disorder stages in the training set as a unified population, where $\mathbb{X}=\{\mathbf{X}_1,\mathbf{X}_2,...,\mathbf{X}_M\}$ in the previous step comprises the nodal feature matrices of all training samples.
Using our SVD-based feature scoring function, we compute a communal feature scoring matrix that quantifies the global importance of each feature to every network node across different disorder stages.  We then compute the average score for each feature to rank the feature dimensions, differentiating between the communal features and the individual features by setting the number of dropping features $\lambda$. We select a subset of features as communal features from both high and low dimensions, namely top-$\lambda$ and bottom-$\lambda$. By dropping these communal features, we achieve denoising and dimensionality reduction of the node features in the brain graph. This results in a total of 2$\lambda$ dropped features, and the denoised node feature matrices $\hat{\mathbb{X}}=\{\mathbf{\hat{X}}_1,\mathbf{\hat{X}}_2,...,\mathbf{\hat{X}}_{M}\}$, where $\hat{\mathbf{X}}_m\in\mathbb{R}^{N\times D^{\prime}}\text{ have reduced dimensionality }D^{\prime}=D-2\lambda$. The communal feature dropping step is shown in Fig. \ref{common}. This process effectively identifies and eliminates noise features in brain graphs that are unrelated to specific disorder stages, thereby enhancing the accuracy of subsequent stage-specific analyses.

\begin{algorithm}
\caption{SVD-based Scoring Function}
\label{algo1}
\begin{algorithmic}[1]
    \STATE \textbf{Input:} Node features of all subjects $\mathbb{X} = \{\mathbf{X}_1, \mathbf{X}_2, \dots, \mathbf{X}_M\}, \mathbb{X}\in\mathbb{R}^{M \times N \times D}$, where $\mathbf{X}_m \in \mathbb{R}^{N \times D}$ is the feature matrix of the $m$-th subject, $N$ is the number of nodes, and $D$ is the number of features.
    \STATE \textbf{Output:} Scoring matrix $\mathbf{S}$
    \STATE Set a node-level tensor $\widetilde{\mathbb{X}}=\{\widetilde{\mathbf{X}}_1,\widetilde{\mathbf{X}}_2,...,\widetilde{\mathbf{X}}_N\},~\widetilde{\mathbb{X}} \in \mathbb{R}^{N \times D \times M}$, where $\tilde{\mathbf{X}}_n \in \mathbb{R}^{D \times M}$ is the the matrix containing node features across all subjects of node $n$.
        \FOR{$i=1$ \TO $M$}
            \FOR{$j=1$ \TO $N$}
                \FOR{$z=1$ \TO $D$}
                    \STATE $\widetilde{\mathbb{X}}[j,z,i] = \mathbb{X}[i,j,z]$
                \ENDFOR
            \ENDFOR
        \ENDFOR
        \FOR{$i=1$ \TO $N$}
            \STATE $\mathbf{S}' = \operatorname{SVD}(\widetilde{\mathbb{X}}[i,:,:])$
            \FOR{$j=1$ \TO $D$}
                \STATE $\mathbf{S}[i,j] = \mathbf{S}'[j,1]$
            \ENDFOR
        \ENDFOR
        \RETURN $\mathbf{S}$
\end{algorithmic}
\end{algorithm}

\subsubsection{Stage Feature Extraction}
% The previous step has performed denoising on the nodal features of the subjects, while stage feature extraction focuses mainly on calculating a node feature scoring matrix for each stage of the disorder.

% In this step, the feature matrices of the denoised node are grouped into $K$ groups, represented as $\hat{\mathbb{X}}=\{\hat{\mathbb{X}}_{k}\}_{k=1}^{K}$. To extract stage features within each group, we again use the SVD-based scoring function for the denoised node feature matrices of all subjects within a single group, obtaining the scoring matrix for each group. For group $k$, the scoring matrix for that group is $\mathbf{F}_{k}=\phi(\hat{\mathbb{X}}_{k})$. After computing for all groups, we can obtain a collection of stage feature scoring matrices $\mathbb{F}=\{\mathbf{F}_{k}\}_{k=1}^{K}$, where $\mathbf{F}_{k}\in\mathbb{R}^{N\times D^{\prime}}$, with each group corresponding to a stage of neurodegenerative disorders. The scoring matrices can be viewed as weights of various nodes on different features within that stage, reflecting the unique information of that stage. Simultaneously, the scoring matrices will be used for graph augmentation, which will be elaborated in the next section.

Building upon the CFD step, we further implement Stage Feature Extraction. Specifically, for each distinct stage of the disorder, we compute a corresponding node feature scoring matrix to quantify the stage-specific contribution of each feature to the nodes. The purified node features obtained after the CFD step are then divided into $K$ groups according to the disorder stage of the subjects, represented as $\hat{\mathbb{X}}=\{\hat{\mathbb{X}}_{k}\}_{k=1}^{K}$. To extract stage features within each group, we again use the SVD-based scoring function for the denoised node feature matrices of all subjects within a single group, obtaining the scoring matrix for each group.

For group $k$, the scoring matrix for that group is $\mathbf{F}_{k}=\phi(\hat{\mathbb{X}}_{k})$. After computing for all groups, we can obtain a collection of stage feature scoring matrices $\mathbb{F}=\{\mathbf{F}_{k}\}_{k=1}^{K}$, where $\mathbf{F}_{k}\in\mathbb{R}^{N\times D^{\prime}}$, with each group corresponding to a stage of neurodegenerative disorders. The scoring matrices can be viewed as weights of various nodes on different features within that stage, reflecting the unique information of that stage. Simultaneously, the scoring matrices will be used for graph augmentation, which will be elaborated in Section~\ref{4.b}.

 \begin{figure}[h]
	\centering
	\includegraphics[width=\columnwidth]{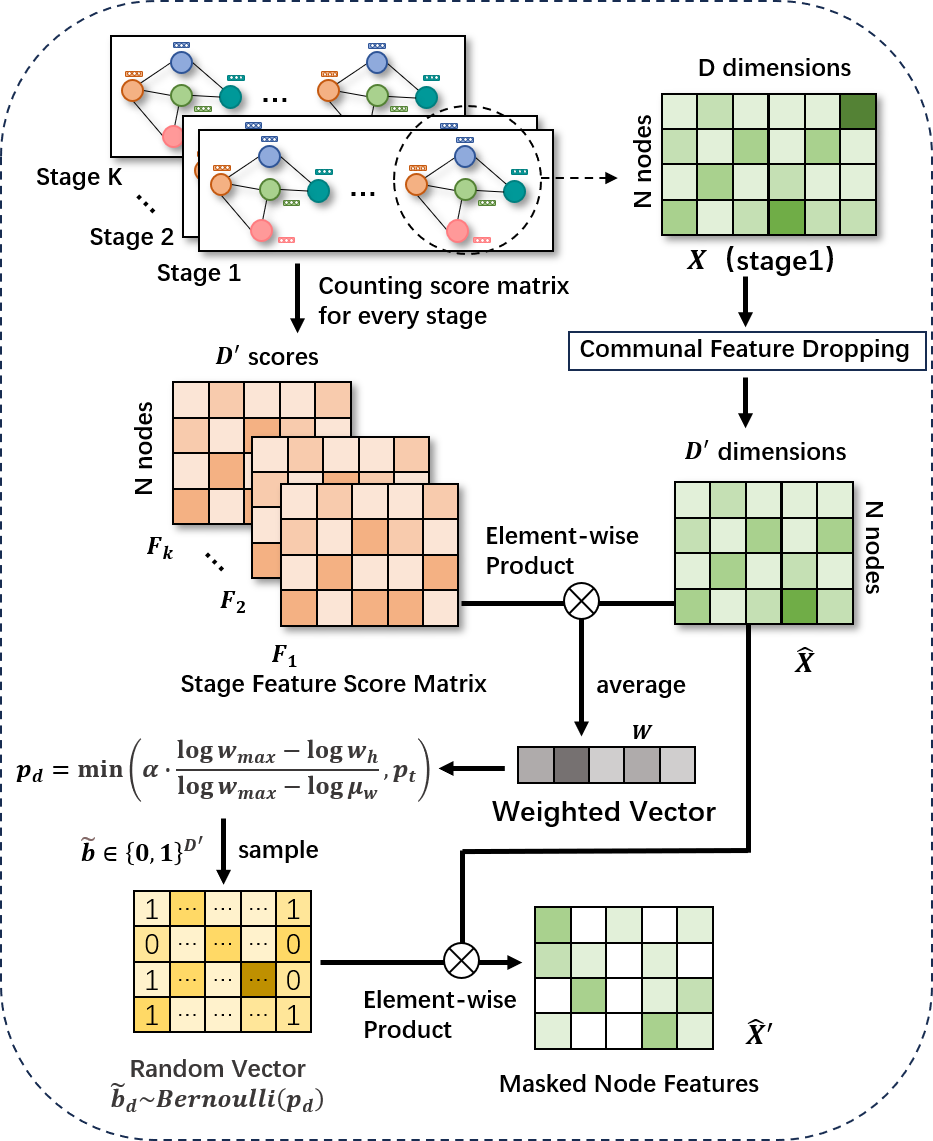}
	\caption{Overview of the proposed stage-adaptive brain graph augmentation approach. We select the brain graph of a single subject from one stage as an example to illustrate how smBGC utilizes the scoring matrix for graph augmentation.}
	\label{augment}
\end{figure}

\subsection{Stage-adaptive Brain Graph Augmentation}\label{4.b}
Graph augmentation has become an indispensable step in graph contrastive learning. The goal of graph augmentation is to generate two graph views by modifying structural information without affecting the semantic labels of the input graphs \cite{liu2024deep}. Instead of relying on uniform data augmentation schemes, such as random node dropping, edge perturbation, or subgraph sampling, we introduce a unique stage feature-based strategy into our graph augmentation scheme. In this section, we will describe how to use  stage feature scoring matrices to augment brain graphs of subjects at different stages. Fig. \ref{augment} illustrates the process of augmentation. 

After applying the stage-aware feature enhancement strategy, we obtain $\hat{\mathbb{X}}^f=\{\hat{\mathbf{X}}_1^f,\hat{\mathbf{X}}_2^f,...,\hat{\mathbf{X}}_M^f\}$ and $\hat{\mathbb{X}}^s=\{\hat{\mathbf{X}}_1^s,\hat{\mathbf{X}}_2^s,...,\hat{\mathbf{X}}_M^s\}$, where $\hat{\mathbb{X}}^f$ and $\hat{\mathbb{X}}^s$ represent the denoised node features of the FBG and SBG, respectively, along with the collection of stage feature scoring matrices  $\mathbb{F}=\{\mathbf{F}_{k}\}_{k=1}^{K}$. Inspired by the GCA method proposed by Zhu \textit{et al.} \cite{zhu2021graph}, which adaptively enhances the edges and node attributes of input graphs, we employ a stage-based graph augmentation module. Our feature masking strategy masks certain less important attributes in the node features. Specifically, we treat stage features as important attributes that should be preserved. For instance, in single-modality graph augmentation, to compute the importance of stage features, we define weights for each dimension of the subject’s node features as:
\begin{eqnarray}
\mathbf{W}=\frac{1}{N}\sum_{i=1}^N||\mathbf{x}_i|\odot \mathbf{F}_{k,i}|,
\end{eqnarray}
where $\mathbf{x}_i$ is the feature vector of node $v_i$, and $\mathbf{F}_{k,i}$ represents the stage scoring vector of node $v_i$ in stage $k$ for the subject. $\odot$ denotes element-wise multiplication, resulting in $\mathbf{W}=[\mathbf{w}_1,\mathbf{w}_2,...,\mathbf{w}_D]$, where $\mathbf{w}_d$ quantifies the aggregated stage-specific importance of the $d$-th feature dimension. To convert these weights into meaningful masking probabilities, we apply a logarithmic transformation to compress their dynamic range and compute the masking probability $p_d$ for each dimension:
\begin{eqnarray}
p_d=\min\left(\alpha\cdot\frac{\log \mathbf{w}_{max}-\log \mathbf{w}_d}{\log \mathbf{w}_{max}-\log\boldsymbol{\mu}_w},p_t\right).
\end{eqnarray}  
where $\mathbf{w}_{max}$ and $\boldsymbol{\mu}_w$ denote the maximum and mean values of $\mathbf{W}$, respectively. Dimensions with higher stage-specific importance (i.e.,$\mathbf{w}_d$ closer to $\mathbf{w}_{max}$) yield smaller $p_d$, making them less likely to be masked. Here, $\alpha$  controls the overall augmentation strength, while $p_t<1$ serves as a truncation threshold to prevent excessively aggressive masking that could disrupt the intrinsic graph structure. The feature enhancement strategy samples a random vector $\tilde{\mathbf{b}}\in\{0,1\}^{D^{\prime}}$, where each dimension is described by a Bernoulli distribution, such that $\tilde{\mathbf{b}}_d{\sim }\textit{Bernoulli}(p_d), \forall i$. Accordingly, $\tilde{\mathbf{b}}_d{\sim }\textit{Bernoulli}(p_d)$ is a masking indicator, where $\tilde{b}_{d}=1$ denotes that the corresponding feature is masked (set to zero) and $\tilde{b}_{d}=0$ denotes that it is retained, and then the enhanced node features are computed as:    
\begin{eqnarray}    \hat{\mathbf{X}}^{\prime}=\left[\hat{\mathbf{x}}_1\odot(1-\tilde{\mathbf{b}}_1);\hat{\mathbf{x}}_2\odot(1-\tilde{\mathbf{b}}_2);...;\hat{\mathbf{x}}_N\odot(1-\tilde{\mathbf{b}}_n)\right]^{\top}.
\end{eqnarray}

Finally, by applying our graph augmentation approach to FBGs and SBGs, we can generate enhanced graph views $\{\hat{G}^{s},\hat{G}^{f}\}$.

\subsection{Multimodal Fusion via Contrastive Learning}

 In this section, we will discuss how to fuse structural and functional brain graphs via contrastive learning and design contrastive loss and optimization objectives for model training.

\subsubsection{Multimodal Brain Graph Embedding}

After the aforementioned steps, we obtain a new augmented graph dataset $\hat{B}=$ $\{(\{\hat{G}_{i}^{s},\hat{G}_{i}^{f}\},y^{i})\}_{i=1}^{M}$, where the node feature matrix of $\hat{G}_i^*$ is $\hat{\mathbf{X}}_i^{*\prime}=[\hat{\mathbf{x}}_1^i,\hat{\mathbf{x}}_2^i,...,\hat{\mathbf{x}}_N^i]^T$. We use a GNN-based view encoder $g(\cdot)$ to compute the embedding for each subject, with the encoder's parameters shared across all modalities. For the embedding of a single-modality brain graph, we input its node features into a message-passing GNN. Inspired by BrainGB \cite{cui2022braingb}, we construct a message vector $\mathbf{v}_{ij}$, which consists of the node embedding $\mathbf{h}_i^{(l)}$ of node $v_i$ at the $l$-th layer, its neighbor's node embedding, and the edge weight $\mathbf{e}_{ij}$, to be fed into an MLP. We aggregate all neighborhood information and perform propagation. Formally, the above process can be represented as:
\begin{eqnarray}    \mathbf{v}_{ij}^{(l)}=MLP\left(\left[\mathbf{h}_i^{(l)};\mathbf{h}_j^{(l)};\mathbf{e}_{ij}\right]\right).
\end{eqnarray}
\begin{eqnarray}
\mathbf{h}_i^{(l)}=ReLU\left(\sum_{j\in N_i\cup\{i\}}\mathbf{v}_{ij}^{(l-1)}\right).
\end{eqnarray}
Finally, we use sum pooling to aggregate all the node embeddings and employ an MLP with residual connections parameterized by $\Phi$ to compute the graph embedding:
\begin{eqnarray}
\mathbf{z}=MLP_\Phi\left(\sum_{i=1}^N \mathbf{h}_i^{(k)}\right)+\sum_{i=1}^N \mathbf{h}_i^{(k)}.
\end{eqnarray}

Through this method, we compute graph-level representations for two modalities, denoted as $\mathbf{z}_i^s{\sim} g(\hat{G}_i^s)$ and $\mathbf{z}_i^f{\sim} g(\hat{G}_i^f)$. %incorporating stage-enhanced information from the brain structure or functional data.

\subsubsection{Multimodal Fusion and Optimization}    

We calculate graph-level representations for each subject, where the graph representations for structural and functional views are represented as $\mathbf{z}_i^s{\sim}g(\hat{G}_i^s)$ and $\mathbf{z}_i^f{\sim}g(\hat{G}_i^f)$. During inference, we fuse the representations of the two views by averaging the sum to form the multimodal graph representation for downstream tasks, expressed as: $\mathbf{z}_i=(\mathbf{z}_i^s+\mathbf{z}_i^f)/2$.

We adopt a global-to-local contrastive learning mode to set an optimization objective for model training. Specifically, we contrast the node representation of one view with the graph representation of another view, aiming to maximize the mutual information between views, defined as:
\begin{gather}
\mathcal{L}_{con}=\notag \\
\frac1{2M}\sum_{(\hat{G}_i^s,\hat{G}_i^f)\in\hat{B}}\left[\frac1N\sum_{j=1}^N\left(I(\mathbf{h}_j^f;\mathbf{z}_i^s)+I(\mathbf{h}_j^s;\mathbf{z}_i^f)\right)\right].
\end{gather}
Here, $I(\mathbf{h}_j^f;\mathbf{z}_i^s)$ estimates the mutual information between the node representation $\mathbf{h}_j^f$ in the functional view and the structural graph representation $\mathbf{z}_i^\mathrm{s}$ of the $i$-th subject, and vice versa. By setting this contrasting objective between different modalities, we ensure cross-modal consistency, allowing the model to adaptively extract discriminative knowledge from each modality without learning a shared brain graph structure. We employ the Jensen-Shannon Divergence (JSD) to compute mutual information $I(\mathbb{X};\mathbb{Y})$. For example, the JSD mutual information estimate for $I(\mathbf{h}_j^f;\mathbf{z}_i^s)$  can be represented as:
\begin{gather}
I\big(\mathbf{h}_j^f;\mathbf{z}_i^s\big)=\notag \\
-sp\left(-d\big(\mathbf{h}_j^f,\mathbf{z}_i^s\big)\right)-\frac{1}{N-1}\sum_{\nu_n\in V\setminus\{\nu_j\}}sp\left(d\big(\mathbf{h}_n^f,\mathbf{z}_i^s\big)\right).
\end{gather}
Here, $sp(x)=\log{(1+e^x)}$ denotes the softplus method, and $d(x,y)$ is a discriminator function that computes the inner product of the two node embeddings along with a sigmoid activation function. The left part of the equation calculates the score for positive sample pairs, taking the functional node embedding of node $v_j$ and its corresponding structural graph embedding as positive samples, while the right part computes the score for negative sample pairs, taking all node embeddings except node $v_j$ and their graph embeddings as negative samples. The process of maximizing mutual information encourages the scores of positive sample pairs to increase while reducing the scores of negative sample pairs. The theoretical optimization objective is to minimize the contrastive loss $\mathcal{L}_{con}$, which corresponds to maximizing the lower bound of the mutual information between positive sample pairs across views. 

Finally, we employ a multilayer perceptron to generate predictions for classification tasks, using cross-entropy loss as the supervised loss. The entire model can be trained in an end-to-end manner by optimizing:
\begin{eqnarray}
\mathcal{L}=\eta_1\mathcal{L}_{con}+\eta_2\mathcal{L}_{supervised},
\end{eqnarray}
where $\eta_{1}$ and $\eta_{2}$ are the loss weights.

\section{Experiments}\label{experiment}

In this section, we conduct a series of experiments to evaluate the effectiveness of the smBGC method. First, we provide detailed information about the datasets and experimental setup. Then, we perform experiments on two datasets to compare the performance of different methods. Finally, we conduct the ablation study and parameter study.

\subsection{Datasets and Preprocessing}

We evaluate our method using two publicly available datasets: the third Alzheimer’s Disease Neuroimaging Initiative (ADNI-3) dataset \cite{jack2008alzheimer}  and the Parkinson’s Progression Markers Initiative (PPMI) dataset \cite{marek2018parkinson}. (1) The ADNI-3 is a longitudinal, multi-site, multimodal neuroimaging dataset. In this study, we focus on the latest ADNI-3 dataset, which includes more comprehensive imaging data across additional disorder stages. We collect data from 578 subjects for evaluation, comprising 54 AD patients, 34 late mild cognitive impairment (LMCI) patients, 192 MCI patients, 88 early mild cognitive impairment (EMCI) patients, and 210 NC. Using the ADNI-3 dataset, we perform a three-stage classification task (NC vs. MCI vs. AD) and a four-stage classification task (NC vs. EMCI vs. LMCI vs. AD). (2) The PPMI dataset includes various types of neuroimaging data across different stages of PD. We collect data from 56 NC, 100 high-risk prodromal patients, and 99 PD patients. Using the PPMI dataset, we perform a three-stage classification task (NC vs. Prodromal vs. PD).

For both datasets, each subject includes both fMRI and DTI data. We exclude duplicate scans and retain single-scan data for each modality. For fMRI data, we process it using the GRETNA Toolbox \cite{wang2015gretna}, segmenting the brain into regions based on the AAL90 template, calculating time series for each region, and measuring functional connectivity using the Pearson correlation coefficient (PCC). For DTI data, we process it using PANDA \cite{cui2013panda}, correct distortions with FSL to obtain fiber images, and segment brain regions using the AAL90 template based on T1-weighted images. Structural connectivity is calculated using fractional anisotropy metrics.

\begin{algorithm}[t]
\caption{Training Procedure of smBGC}
\label{alg:train}
\begin{algorithmic}[1]
\REQUIRE Training set $\mathcal{B}=\{(\{G_i^s,G_i^f\},y_i)\}_{i=1}^{M}$; number of stages $K$; number of dropping features $\lambda$; loss weights $\eta_1,\eta_2$
\ENSURE Trained encoder $g(\cdot)$, classifier, dropped feature index set $\mathcal{I}_{\text{drop}}$
\STATE Compute communal scoring matrix $S \leftarrow \phi(\mathbf{X})$ over all training subjects (Eq. 1--3)
\STATE Rank features by average score; select top-$\lambda$ and bottom-$\lambda$ as $\mathcal{I}_{\text{drop}}$
\STATE Drop $\mathcal{I}_{\text{drop}}$ from all node features to obtain $\hat{\mathbf{X}}$ \COMMENT{Communal Feature Dropping}
\STATE Partition $\hat{\mathbf{X}}$ into $K$ stage groups $\{\hat{\mathbf{X}}_k\}_{k=1}^{K}$
\FOR{$k=1$ \TO $K$}
    \STATE Compute stage scoring matrix $F_k \leftarrow \phi(\hat{\mathbf{X}}_k)$ \COMMENT{Stage Feature Extraction}
\ENDFOR
\FOR{epoch $=1$ \TO $E$}
    \FOR{each subject $i$ in a mini-batch}
        \STATE Compute importance weights $\mathbf{W}$ via Eq. 6 using $F_{y_i}$
        \STATE Compute masking probabilities $p_d$ via Eq. 7
        \STATE Sample masking indicators $\tilde{b}_d\sim\text{Bernoulli}(p_d)$
        \STATE Obtain augmented views $\hat{G}_i^s,\hat{G}_i^f$ via Eq. 8
        \STATE Encode both views with shared GNN encoder $g(\cdot)$ to get $z_i^s,z_i^f$ and node embeddings
        \STATE Compute contrastive loss $\mathcal{L}_{con}$ (Eq. 12--13) between the two views
    \ENDFOR
    \STATE Fuse embeddings $z_i=(z_i^s+z_i^f)/2$; compute classification loss $\mathcal{L}_{supervised}$
    \STATE Update parameters by minimizing $\mathcal{L}=\eta_1\mathcal{L}_{con}+\eta_2\mathcal{L}_{supervised}$ (Eq. 14)
\ENDFOR
\RETURN trained $g(\cdot)$, classifier, $\mathcal{I}_{\text{drop}}$
\end{algorithmic}
\end{algorithm}

\begin{algorithm}[t]
\caption{Testing Procedure of smBGC}
\label{alg:test}
\begin{algorithmic}[1]
\REQUIRE Test subject with $\{G^s,G^f\}$; trained encoder $g(\cdot)$; trained classifier; dropped feature index set $\mathcal{I}_{\text{drop}}$ (fixed from training)
\ENSURE Predicted disorder stage $\hat{y}$
\STATE Remove features indexed by $\mathcal{I}_{\text{drop}}$ from $X^s,X^f$ \COMMENT{no re-computation; mask is fixed}
\STATE Encode $G^s,G^f$ with trained $g(\cdot)$ (no augmentation) to obtain $z^s,z^f$
\STATE Fuse: $z=(z^s+z^f)/2$
\STATE $\hat{y}\leftarrow$ classifier$(z)$
\RETURN $\hat{y}$
\end{algorithmic}
\end{algorithm}

\subsection{Experimental Settings}

To thoroughly evaluate the performance of our method, we use five evaluation metrics, including classification accuracy (ACC), sensitivity (SEN), specificity (SPE), F1 score (F1) and ROC-AUC (AUC), where higher values indicate better performance. In this study, we use 5-fold cross-validation to evaluate all methods. The splits of the training and testing sets remain consistent, with the dataset divided into five subsets. For each experiment, four subsets are used as the training set, and the remaining subset is used as the testing set. 
In the stage-aware feature enhancement strategy, both CFD and SFE are performed exclusively on the training set to compute the noise features and stage-specific scoring matrices. The enhanced training data are then used to train the contrastive learning model. During inference, a fixed feature mask learned from the training set is applied to remove noise features from the test samples, and the processed test data are subsequently fed into the trained model for classification.
All experiments are implemented in PyTorch and conducted on NVIDIA GeForce A40 GPU with 24 GB memory. The hyperparameters used in our model for different stage diagnosis tasks are detailed in Table \ref{tab:settings}. All hyperparameters are selected based on preliminary experiments on the training data and are fixed across all folds. During training, we employ the Adam optimizer, and the model's performance on the testing set is evaluated every five epochs. For each fold in the 5-fold cross-validation, the performance on the testing set at the final epoch is recorded across five runs. The means and standard deviations of these runs are calculated to represent the final results for each metric. The complete training and testing pipeline are illustrated in Algorithm \ref{alg:train} and Algorithm \ref{alg:test}.

\begin{table}[H]
	\centering
	\caption{Hyperparameters for training on three different tasks.}
 \setlength{\tabcolsep}{3pt} % Adjust column separation
    \resizebox{0.5\textwidth}{!}{
\begin{tabular}{c|ccc}
\toprule
Hyperparameter & ADNI-3 (three stages)& ADNI-3 (four stages)& PPMI  \\ \midrule
Functional magnitude $\alpha_f$  & 0.6  & 0.6  & 0.7 \\
Structural magnitude $\alpha_s$  & 0.6  & 0.5  & 0.3 \\
Dropping features $\lambda$  & 20  & 20  & 30 \\
GNN layers    & 1  & 1  & 1 \\
Dropout   & 0.5  & 0.5  & 0.5 \\
Readout method   & sum  & sum  & sum \\
Learning rate    & 1\text{e-}2  & 1\text{e-}2  & 1\text{e-}2 \\
Batch size    & 16  & 16  & 16 \\
Epochs    & 100  & 100  & 100 \\
Weight decay    & 1\text{e-}4  & 1\text{e-}4  & 1\text{e-}4 \\
\bottomrule
\end{tabular}}
\label{tab:settings}
\end{table}

\subsection{Model Comparison}

\subsubsection{Baselines}
In this study, to evaluate the effectiveness of smBGC, we compare it with deep graph learning models and well-established brain graph learning models. These methods include:
\begin{itemize}
    \item \textbf{GCN} \cite{kipf2017semisupervised}: Graph Convolutional Network (GCN) performs convolution operations on the graph in either the spectral or spatial domain, leveraging the inherent graph structure.
    \item \textbf{GAT} \cite{DBLP:conf/iclr/VelickovicCCRLB18}: Graph Attention Network (GAT) is a type of graph neural network that uses attention mechanisms to prioritize the influence of neighboring nodes during information aggregation. This mechanism allows GAT to capture complex relationships and dependencies more effectively. 
    \item \textbf{GraphSAGE} \cite{DBLP:conf/nips/HamiltonYL17}: GraphSAGE samples a fixed-sized neighborhood around each node and aggregates the features of these neighbors from large-scale graphs to learn node embeddings.
    \item \textbf{GIN} \cite{DBLP:conf/iclr/XuHLJ19}: Graph Isomorphism Network (GIN) is inspired by the Weisfeiler-Lehman test and aims to provide equivalent discriminative power through expressive aggregation functions.
    \item \textbf{GCA} \cite{zhu2021graph}: GCA is a general contrastive framework for unsupervised graph representation learning with adaptive data augmentation. It performs data augmentation on both topological and attribute levels to adapt to graph structures and attributes, encouraging the model to learn important features from these aspects.
    \item \textbf{BrainGB} \cite{cui2022braingb}: BrainGB is a benchmark for brain graph analysis using GNNs. It standardizes this process and summarizes the pipeline for constructing brain graphs from functional and structural neuroimaging patterns, modularizing the implementation of GNN designs.
    \item \textbf{BrainIB}~\cite{10680255}: BrainIB takes brain networks constructed from fMRI data as input and learns to identify the most discriminative subgraph of key connections, thereby reducing redundant information while improving generalization performance. 
    \item \textbf{MAHGCN}~\cite{10155767}: MAHGCN constructs functional connectivity networks (FCNs) across multiple spatial scales and leverages hierarchical relationships among brain regions at different scales through an Atlas-guided Pooling strategy. By integrating multiscale information within a hierarchical GCN architecture, it is able to more comprehensively extract brain connectivity features. 
    \item \textbf{Cross-GNN}~\cite{yang2023mapping}: Cross-GNN models cross-modal relationships through dynamic brain graphs and mutual learning mechanisms, using data-driven attention modeling on the adjacency matrices that represent modality dependencies. 
    \item \textbf{RH-BrainFS}~\cite{ye2024rh}: RH-BrainFS proposes a strategy for region heterogeneity in multimodal brain graph fusion, integrating heterogeneous region features from different modalities within brain graphs while addressing regional heterogeneity issues.
    \item \textbf{MCGAN}~\cite{9947312}: MCGAN introduces a multioutput conditional GAN for data augmentation, followed by a multiheaded gating fusion model to capture complementary features across modalities for classification. 
    \item \textbf{MMP-GCN}~\cite{song2022multicenter}: MMP-GCN introduces a penalty mechanism based on DTI intensities to construct functional brain graphs, enhancing the expression of connectivity information. This method integrates various data sources and assigns unique filters to different data, thereby enhancing model performance.
\end{itemize}

\subsubsection{Result Comparison}

\begin{table*}[!ht]
	\centering
	\caption{Performance comparison of different methods on the ADNI-3 dataset (\%). The best results are highlighted in bold and the second-best results are underlined.}
	\label{ADNI}
 \setlength{\tabcolsep}{3pt}
 \ssmall
  \resizebox{\textwidth}{!}{
\begin{tabular}{ccccccc|ccccc}
\toprule
\multicolumn{2}{c}{\multirow{2}*{Method}}& \multicolumn{5}{c}{NC vs. MCI vs. AD} & \multicolumn{5}{c}{NC vs. EMCI  vs. LMCI vs. AD} \\\cmidrule{3-12}
& & ACC & F1 & AUC & SEN & SPE & ACC & F1 & AUC  & SEN  & SPE \\
\midrule
& GCN         & 54.30 $\pm$ 1.67 & 41.97 $\pm$ 2.29 & 67.51 $\pm$ 2.72 & 42.80 $\pm$ 1.79 & 74.06 $\pm$ 1.19 & 43.75 $\pm$ 3.64 & 35.01 $\pm$ 3.60 & 65.90 $\pm$ 3.58 & 35.07 $\pm$ 3.60 & 80.83 $\pm$ 1.29         \\
& GAT         & 50.78 $\pm$ 2.44         & 41.65 $\pm$ 3.48         & 65.22 $\pm$ 2.99         & 41.79 $\pm$ 2.86         & 72.07 $\pm$ 0.88 & 46.88 $\pm$ 7.23 & 37.07 $\pm$ 5.31 & 69.95 $\pm$ 5.03 & 39.17 $\pm$ 4.77 & 81.38 $\pm$ 1.87        \\
& GIN         & 50.77 $\pm$ 2.57         & 42.38 $\pm$ 2.48         & 63.06 $\pm$ 2.45         & 43.19 $\pm$ 2.75         & 72.89 $\pm$ 1.42 & 50.00 $\pm$ 11.63 & 47.28 $\pm$ 11.58 & 63.51 $\pm$ 7.34 & 47.22 $\pm$ 10.95 & 82.56 $\pm$ 3.93        \\
& GraphSAGE   & 56.95 $\pm$ 4.13         & 47.74 $\pm$ 5.15         & 69.47 $\pm$ 3.56         & 47.84 $\pm$ 5.02         & 75.55 $\pm$ 2.87 & 43.75 $\pm$ 6.12 & 39.29 $\pm$ 5.14 & 61.55 $\pm$ 4.87 & 40.84 $\pm$ 4.46 & 81.26 $\pm$ 2.47        \\
& BrainGB     & 53.33 $\pm$ 4.46         & 38.23 $\pm$ 4.79         & 62.10 $\pm$ 3.09         & 40.84 $\pm$ 4.12         & 72.35 $\pm$ 2.81 & 40.63 $\pm$ 5.92 & 32.14 $\pm$ 6.02 & 55.68 $\pm$ 4.01 & 34.67 $\pm$ 4.20 & 76.02 $\pm$ 1.79        \\
\midrule
& MAHGCN  &56.80 $\pm$ 4.28        & 42.04 $\pm$ 4.12        & 71.33 $\pm$ 5.08         & 44.60 $\pm$ 3.48         & 74.58 $\pm$ 2.22  & 41.95 $\pm$ 5.43 & 32.36 $\pm$ 3.79 & 68.27 $\pm$ 2.35 & 36.79 $\pm$ 4.84 & 79.85 $\pm$ 1.79       \\
& BrainIB  &61.62 $\pm$ 1.40        & 36.88 $\pm$ 5.21         & 66.59 $\pm$ 2.67         & 48.38 $\pm$ 3.50         & 76.89 $\pm$ 0.79  & 49.76 $\pm$ 5.89 & 41.51 $\pm$ 6.71 & 62.65 $\pm$ 2.10 & 44.23 $\pm$ 5.97 & 82.40 $\pm$ 2.08       \\

& RH-BrainFS  & 57.14 $\pm$ 2.05         & 50.24 $\pm$ 1.51         & 61.30 $\pm$ 1.68         & 42.52 $\pm$ 1.47         & 73.83 $\pm$ 1.16  & 47.62 $\pm$ 6.70 & 48.33 $\pm$ 7.57 & 66.48 $\pm$ 6.10 & 40.00 $\pm$ 6.46 & 66.67 $\pm$ 9.14       \\
& MMP-GCN     & 63.22 $\pm$ 2.64         & 55.48 $\pm$ 2.21         & 70.48 $\pm$ 2.18         & 53.27 $\pm$ 1.74         & 75.79 $\pm$ 2.16   & 41.03 $\pm$ 6.47 & 36.36 $\pm$ 2.34 & 60.09 $\pm$ 7.85 & 36.39 $\pm$ 4.83 & 79.51 $\pm$ 1.95       \\
& Cross-GNN   & 64.24 $\pm$ 1.37         & 45.49 $\pm$ 0.94         & 77.53 $\pm$ 0.53         & 42.77 $\pm$ 0.79         & 80.10 $\pm$ 0.84  & 40.63 $\pm$ 2.34 & 28.79 $\pm$ 2.16 & 64.12 $\pm$ 3.15 & 31.11 $\pm$ 3.17 & 79.36 $\pm$ 1.34 \\
& MCGAN  &60.96 $\pm$ 3.42        & 57.44 $\pm$ 5.24         & 71.29 $\pm$ 3.31        & 50.15 $\pm$ 8.64         & 79.37 $\pm$ 2.23  & \underline{57.56 $\pm$ 4.20} & 47.41 $\pm$ 2.63 & 69.98 $\pm$ 5.06 & \underline{50.78 $\pm$ 2.92} & \underline{85.23 $\pm$ 1.15}       \\
& GCA         & \underline{69.34 $\pm$ 3.98} & \underline{59.21 $\pm$ 4.80} & \underline{85.25 $\pm$ 6.41} & \underline{56.83 $\pm$ 3.84} & \underline{81.32 $\pm$ 2.53}  & 52.03 $\pm$ 8.57 & \underline{48.12 $\pm$ 7.05} & \underline{74.75 $\pm$ 1.16} & 48.48 $\pm$ 7.50 & 83.65 $\pm$ 2.83     \\ 
\midrule
& \textbf{smBGC (ours)} & \textbf{96.47 $\pm$ 2.32} & \textbf{96.18 $\pm$ 2.53} & \textbf{99.06 $\pm$ 0.62} & \textbf{95.45 $\pm$ 2.60} &\textbf{97.89 $\pm$ 1.38} & \textbf{81.71 $\pm$ 1.22} & \textbf{79.02 $\pm$ 0.56} & \textbf{96.14 $\pm$ 0.43} & \textbf{64.97 $\pm$ 2.27} & \textbf{96.24 $\pm$ 0.53}  \\ \bottomrule
\end{tabular}}
\end{table*}

\begin{table}[ht]
	\centering
	\caption{Performance comparison of different methods on the PPMI dataset (\%).}
	\label{PPMI}
 \setlength{\tabcolsep}{3pt} % Adjust column separation	
 \ssmall
  \resizebox{0.5\textwidth}{!}{
\begin{tabular}{ccccccc}
\toprule
\multicolumn{2}{c}{\multirow{2}*{Method}}& \multicolumn{5}{c}{NC vs. Prodromal vs. PD}  \\\cmidrule{3-7}
& & ACC & F1 & AUC & SEN & SPE  \\
\midrule
& GCN         & 51.92 $\pm$ 3.57 & 49.55 $\pm$ 3.19 & 66.26 $\pm$ 4.91 & 50.34 $\pm$ 2.23 & 74.66 $\pm$ 0.66         \\
& GAT         & 50.58 $\pm$ 5.66 & 46.22 $\pm$ 1.68 & 60.32 $\pm$ 3.08 & 48.00 $\pm$ 1.62 & 73.20 $\pm$ 5.56       \\
& GIN         & 50.52 $\pm$ 5.10 & 49.24 $\pm$ 4.23 & 68.13 $\pm$ 3.94 & 49.83 $\pm$ 4.42 & 74.25 $\pm$ 2.43      \\
& GraphSAGE   & 55.04 $\pm$ 2.67 & 52.46 $\pm$ 5.20 & 66.91 $\pm$ 5.77 & 53.22 $\pm$ 4.35 & 76.19 $\pm$ 1.29       \\
& BrainGB     & 48.10 $\pm$ 5.39 & 42.80 $\pm$ 4.26 & 60.82 $\pm$ 2.62 & 45.51 $\pm$ 2.99 & 72.29 $\pm$ 1.75     \\
\midrule
& MAHGCN  & 43.59 $\pm$ 6.82 & 30.78 $\pm$ 8.30 & 68.21 $\pm$ 8.46 & 39.55 $\pm$ 4.15 & 69.74 $\pm$ 1.95   \\
& BrainIB  & 52.54 $\pm$ 2.56 & 47.47 $\pm$ 4.75 & 59.27 $\pm$ 4.10 & 51.09 $\pm$ 3.23 & 74.88 $\pm$ 1.45   \\
& RH-BrainFS  & 58.38 $\pm$ 1.58 & 54.88 $\pm$ 2.11 & 67.30 $\pm$ 2.92 & 47.03 $\pm$ 1.08 & 78.62 $\pm$ 0.65   \\
& MMP-GCN     & 60.06 $\pm$ 2.47 & 55.03 $\pm$ 3.58 & 64.61 $\pm$ 3.77 & 55.95 $\pm$ 5.81 & 79.55 $\pm$ 1.81      \\
& Cross-GNN   & 56.60 $\pm$ 1.89 & 48.25 $\pm$ 1.51 & 70.14 $\pm$ 3.37 & 49.64 $\pm$ 1.35 & 79.64 $\pm$ 1.32  \\
& MCGAN  & \underline{66.41 $\pm$ 5.41} & \underline{64.30 $\pm$ 5.68} & 70.12 $\pm$ 3.78 & \underline{62.38 $\pm$ 5.51} & \underline{84.75 $\pm$ 3.45}   \\
& GCA         &65.89 $\pm$ 6.43 & 61.75 $\pm$ 5.89 & \underline{80.39 $\pm$ 6.10} & 61.64 $\pm$ 7.91 & 81.24$\pm$ 3.97       \\
 \midrule
& \textbf{smBGC (ours)} & \textbf{91.15 $\pm$ 3.19 } & \textbf{91.24 $\pm$ 3.13} & \textbf{96.27 $\pm$ 0.91} & \textbf{91.77 $\pm$ 4.33} & \textbf{95.61 $\pm$ 3.52}  \\ \bottomrule
\end{tabular}}
\end{table}

\begin{table}[ht]
	\centering
	\caption{Fold-level performance of smBGC on the four-stage classification task of the ADNI-3 dataset (\%).}
	\label{fold}
 \setlength{\tabcolsep}{3pt} % Adjust column separation	
 \ssmall
  \resizebox{0.5\textwidth}{!}{
\begin{tabular}{lccccccc}
\toprule
Metric
& Fold 1
& Fold 2
& Fold 3
& Fold 4
& Fold 5
& Mean $\pm$ Std.
& 95\% CI \\
\midrule
ACC
& 80.12
& 82.68
& 82.90
& 82.11
& 80.74
& $81.71 \pm 1.22$
& [80.19, 83.23] \\

F1
& 78.43
& 79.12
& 79.78
& 79.25
& 78.52
& $79.02 \pm 0.56$
& [78.33, 79.71] \\

AUC
& 96.55
& 96.28
& 95.73
& 96.51
& 95.63
& $96.14 \pm 0.43$
& [95.60, 96.68] \\

SEN
& 62.42
& 67.21
& 63.85
& 67.54
& 63.83
& $64.97 \pm 2.27$
& [62.15, 67.79] \\

SPE
& 96.72
& 95.81
& 96.65
& 95.54
& 96.48
& $96.24 \pm 0.53$
& [95.58, 96.90] \\
\bottomrule
\end{tabular}
}
\end{table}

\begin{table}[ht]
	\centering
	\caption{Two-sided paired t-tests results comparing smBGC 
with MHGF on the ADNI-3 four-stage classification task.}
	\label{t-test}
 \setlength{\tabcolsep}{3pt} % Adjust column separation	
 \ssmall
  \resizebox{0.5\textwidth}{!}{
\begin{tabular}{lcccccc}
\toprule
Metric
& smBGC (\%)
& MHGF (\%)
& Improvement
& 95\% CI
& $t(4)$
& Adjusted $p$ \\
\midrule
ACC & 81.71 & 57.56 & 24.15 & [20.71, 27.59] & 19.47 & $<0.001$ \\
F1  & 79.02 & 47.41 & 31.61 & [27.84, 35.37] & 23.29 & $<0.001$ \\
AUC & 96.14 & 69.98 & 26.16 & [19.65, 32.67] & 11.16 & $<0.001$ \\
SEN & 64.97 & 50.78 & 14.19 & [11.50, 16.87] & 14.65 & $<0.001$ \\
SPE & 96.24 & 85.23 & 11.01 & [9.75, 12.28] & 24.22 & $<0.001$ \\
\bottomrule
\end{tabular}
}
\end{table}

Table~\ref{ADNI} presents the performance of our method on three-stage and four-stage classification tasks using the ADNI-3 dataset. For the NC vs. MCI vs. AD classification task, smBGC achieves the best performance among all methods, with an accuracy of 96.47\%, an F1 score of 96.18\%, an AUC of 99.06\%, a specificity of 97.89\%, and a sensitivity of 95.45\%. Similarly, in the more challenging four-stage classification task (NC vs. EMCI vs. LMCI vs. AD), smBGC maintains outstanding performance, achieving the highest accuracy of 81.71\%, an F1 score of 79.02\%, an AUC of 96.14\%, a sensitivity of 64.97\%, and a specificity of 96.24\%. Table~\ref{PPMI} demonstrates the results of the three-stage classification task (NC vs. Prodromal vs. PD) on the PPMI dataset. Our method achieves the highest accuracy of 91.15\%, an F1 score of 91.24\%, an AUC of 96.27\%, a sensitivity of 91.77\%, and a specificity of 95.61\%, underscoring the robustness and generalizability of our approach.

Overall, basic deep graph learning models, such as GCN, perform poorly across all tasks. These methods, relying solely on unimodal data, fail to effectively capture the characteristics of brain graphs at different disorder stages. In contrast, models like RH-BrainFS, MMP-GCN, and Cross-GNN, which integrate multimodal data, further improve stage classification performance, with complementary information between modalities playing a critical role. Notably, GCA incorporates multimodal data into a contrastive framework for graph augmentation, outperforming all baseline methods across tasks. This demonstrates that contrastive frameworks are also effective for learning multimodal data, and that the graph augmentation module significantly contributes to performance improvements by mitigating noise and enhancing graph attributes. Our method fully leverages complementary information across modalities and considers stage-specific brain dysfunction characteristics. By addressing these factors, we develop a solution that significantly enhances stage classification performance, achieving substantial performance improvements.

\begin{figure*}[htp]
	\centering
	\begin{minipage}[t]{0.25\textwidth}
	\centering\includegraphics[width=1.0\linewidth]{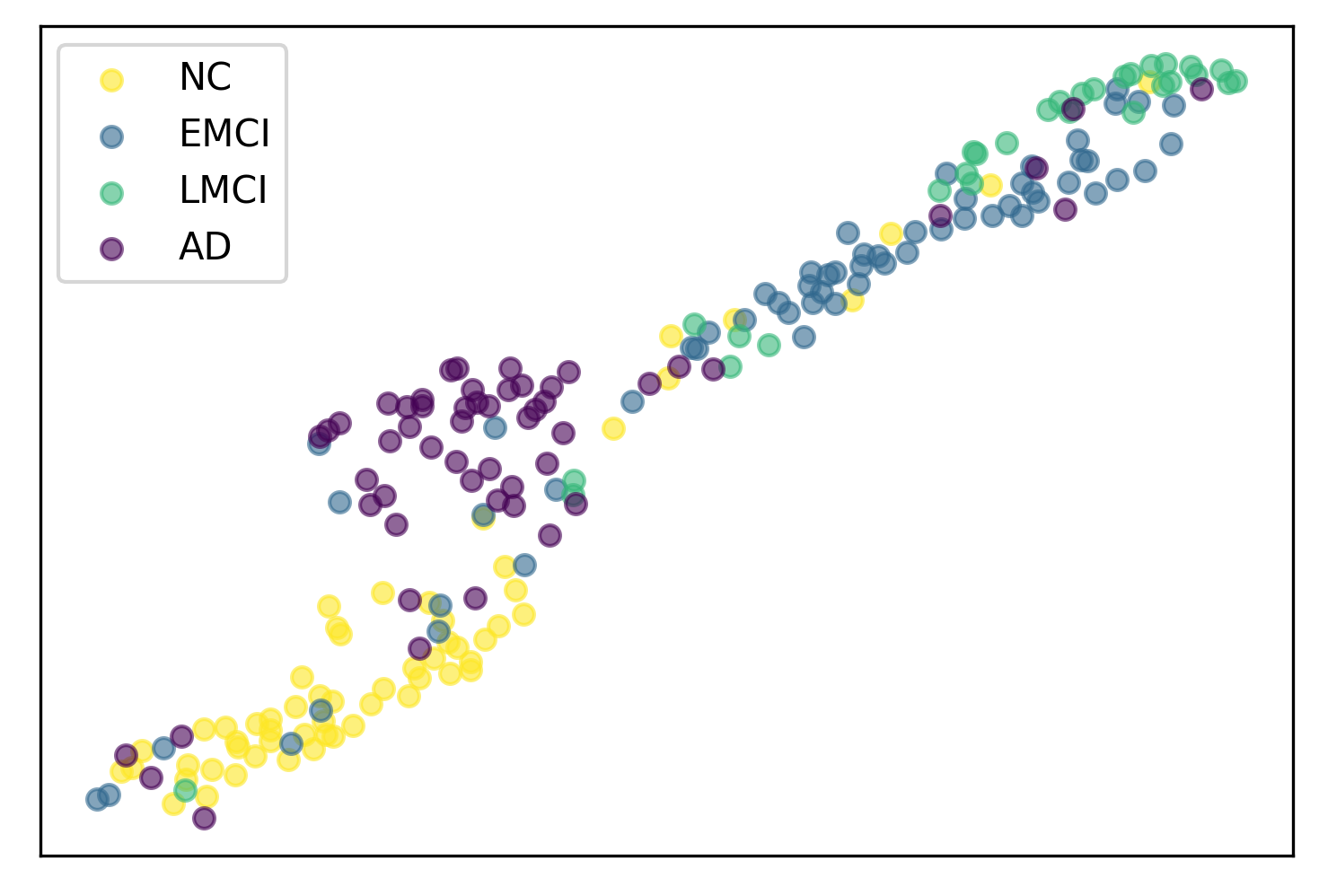}
    \caption*{GraphSAGE}
	\end{minipage}%
	\begin{minipage}[t]{0.25\textwidth}
	\centering\includegraphics[width=1.0\linewidth]{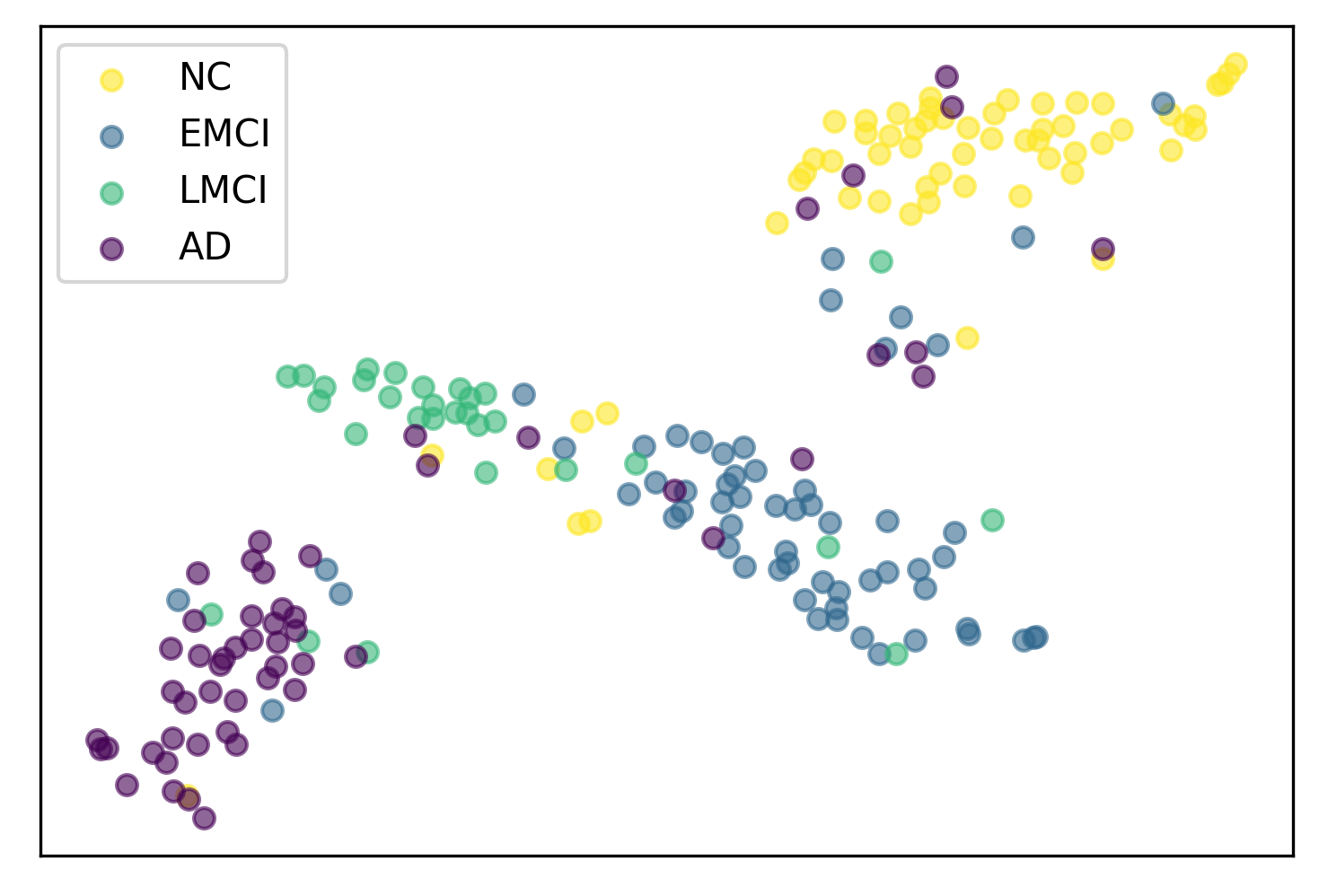}
    \caption*{GCN}
	\end{minipage}%
	\begin{minipage}[t]{0.25\textwidth}
	\centering\includegraphics[width=1.0\linewidth]{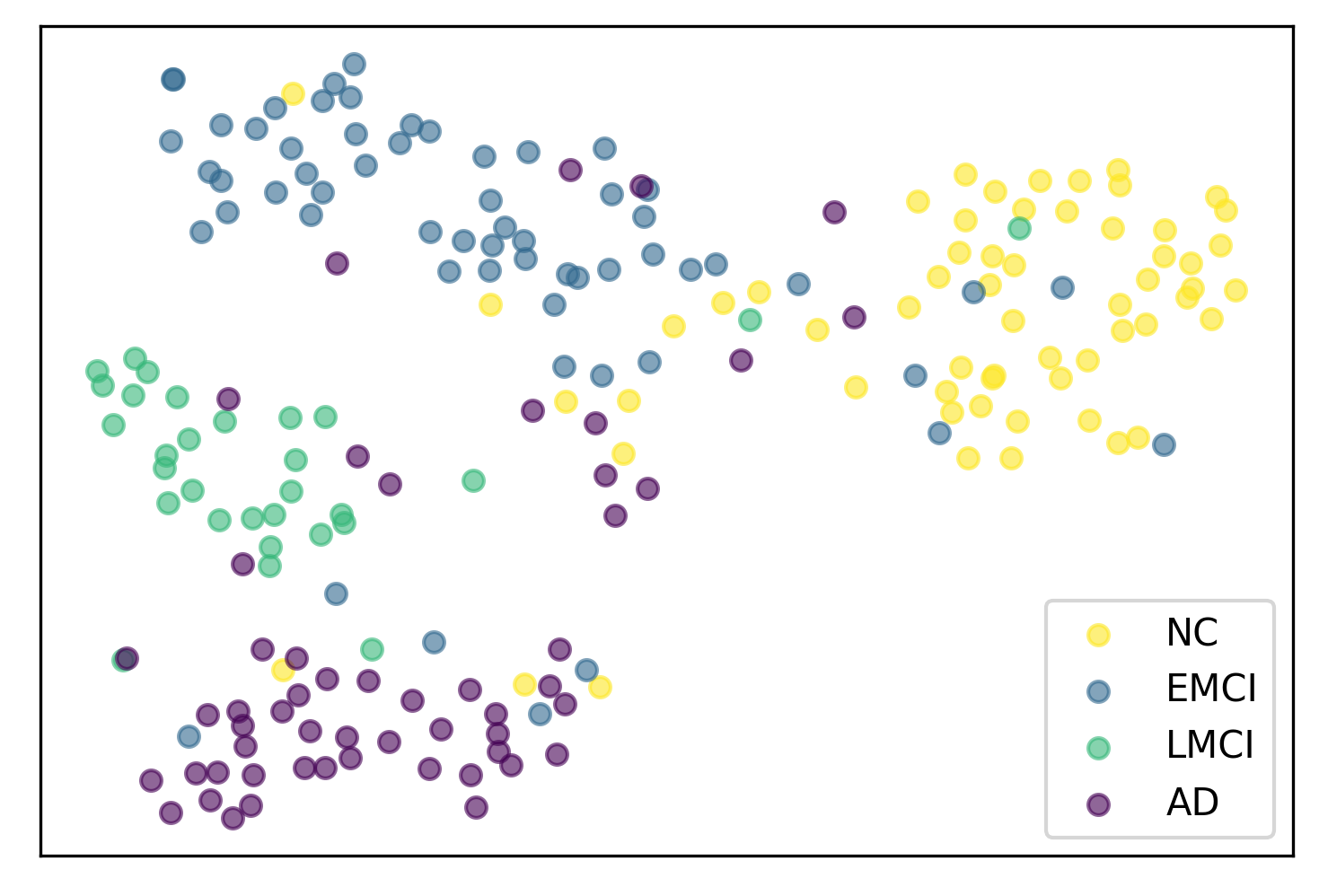}
    \caption*{GAT}
	\end{minipage}%
	\begin{minipage}[t]{0.25\textwidth}
	\centering\includegraphics[width=1.0\linewidth]{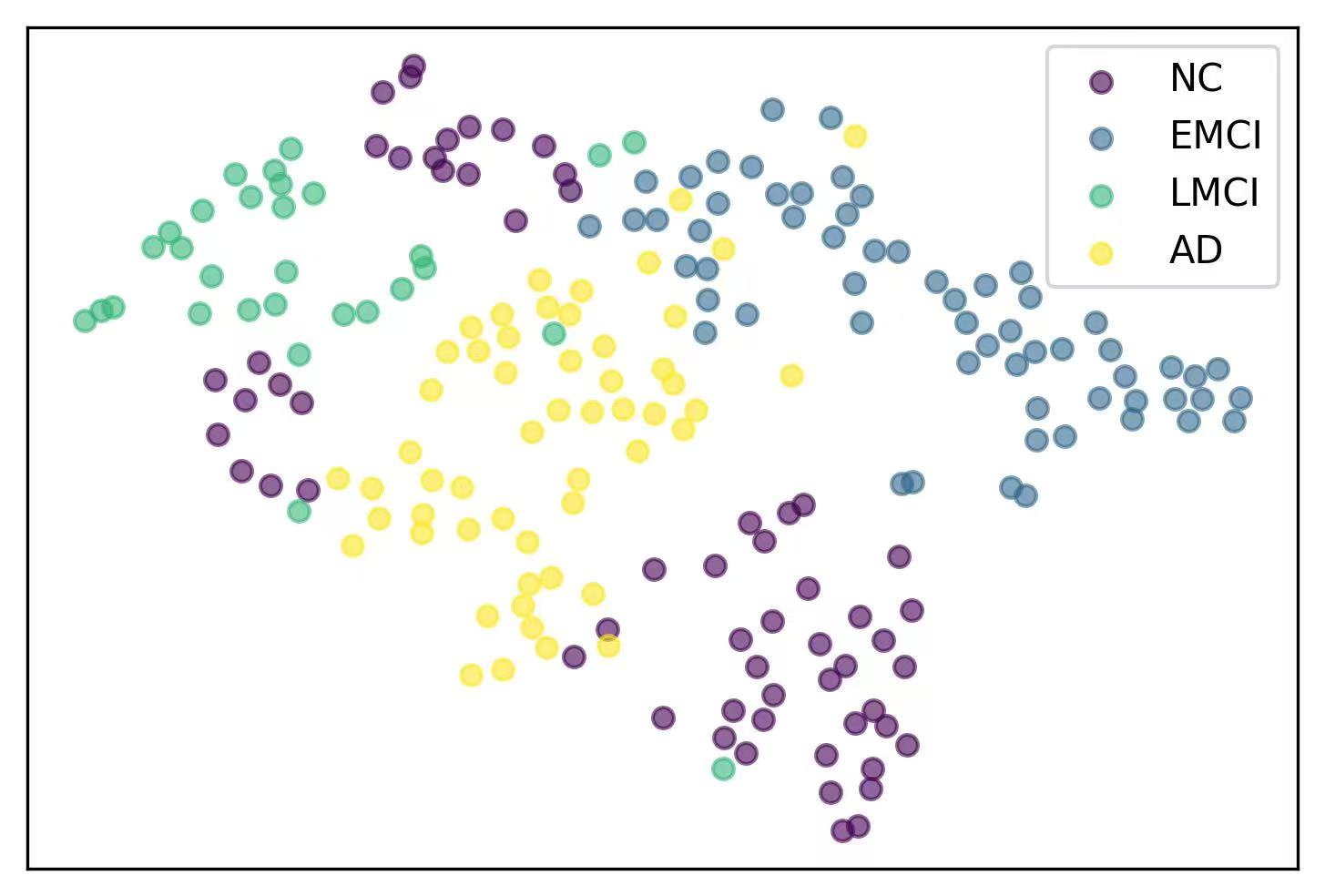}
    \caption*{smBGC}
	\end{minipage}%
	\caption{The t-SNE visualizations of different methods on the ADNI-3 dataset with four stages.}
  \label{fig:tsne}
\end{figure*}

\subsubsection{Visualization}
To demonstrate the performance of various methods, we conduct visualization experiments on the ADNI-3 dataset. Specifically, we use the graph-level embeddings from the final layer of each method for t-SNE visualization. As shown in Fig.~\ref{fig:tsne}, the distribution of GraphSAGE is sparse and fails to effectively separate the four classes of samples. GAT and GCN partially concentrate certain categories into clusters, but these clusters lack well-defined boundaries and exhibit significant overlap. In contrast, our method not only achieves clear clustering but also minimizes confusion at cluster boundaries, with only a small number of misclassified samples.

\subsubsection{Statistical Significance}
To evaluate the robustness of smBGC across different data partitions, we report its fold-level performance and the corresponding 95\% confidence intervals in Table \ref{fold}. The results show that smBGC achieves stable and strong performance across the five cross-validation folds. Specifically, smBGC achieves an average ACC of 81.71\%, an F1 score of 79.02\%, an AUC of 96.14\%, a sensitivity of 64.97\%, and a specificity of 96.24\%. The relatively narrow confidence intervals for ACC, F1, AUC, and specificity indicate that the performance of smBGC is stable across different data partitions.
We further conducted two-sided paired t-tests between smBGC and MHGF using the results obtained from the same five cross-validation folds. As reported in Table \ref{t-test}, smBGC achieves mean improvements of 24.15, 31.61, 26.16, 14.19, and 11.01 percentage points in ACC, F1, AUC, sensitivity, and specificity, respectively. The 95\% confidence intervals of all paired differences are entirely above zero, indicating consistent improvements across the five folds. Moreover, all differences remain statistically significant after applying Holm correction for multiple comparisons (all adjusted p-values $<$0.001).

\begin{figure*}[h!]
\centering
\subfloat[\scriptsize ADNI-3 (three stages)]{
 \begin{minipage}[t]{0.32\textwidth}
	\includegraphics[width=6cm]{./photos/modal_adni3.png}% This is a *.eps file
\end{minipage}}
\subfloat[\scriptsize ADNI-3 (four stages)]{
 \begin{minipage}[t]{0.32\textwidth}
	\includegraphics[width=6cm]{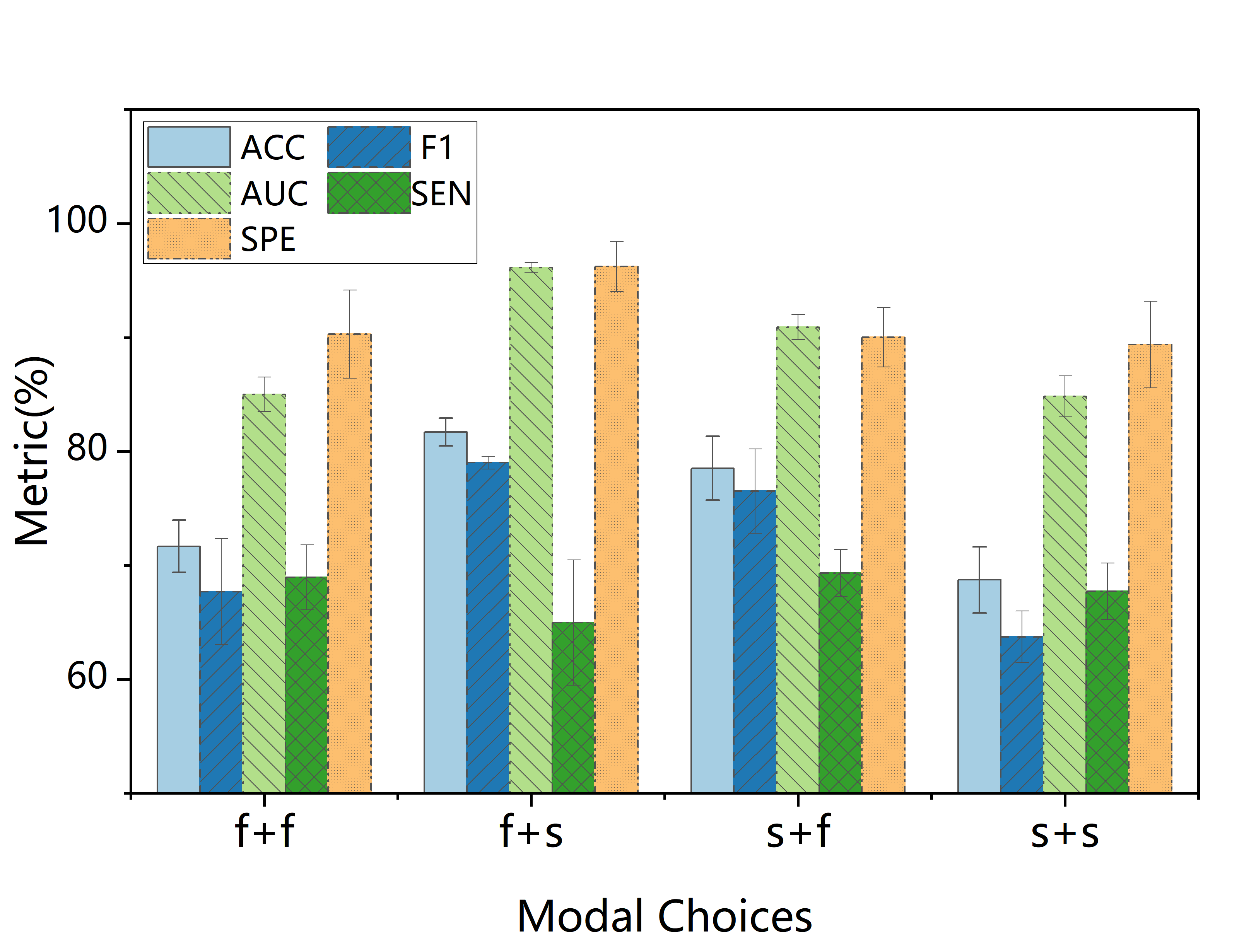}% This is a *.eps file
\end{minipage}}
\subfloat[\scriptsize PPMI]{
 \begin{minipage}[t]{0.32\textwidth}
	\includegraphics[width=6cm]{./photos/modal_ppmi}% This is a *.eps file
\end{minipage}}
	\caption{Experimental results under different modal choices.}
	\label{fig:modal}
\end{figure*}

\subsection{Ablation Study}

In this subsection, we examine the impact of the selected modalities and different modules on the model's performance.

\subsubsection{Analysis of Different Modules}
We employ two modules to obtain stage-specific brain graph features: CFD and graph augmentation based on the results of SFE (ASFE). We validate the effectiveness of both modules. We conduct ablation experiments on the PPMI dataset by removing these modules. As shown in Table \ref{ablation} , a comparison of the first three rows reveals that both the CFD and ASFE modules have a positive impact on our model's performance. Specifically, the comparison between the results of w/o CFD and w/o ASFE highlights the crucial role of the stage-adaptive brain graph augmentation approach in enhancing the model performance. Overall, the results demonstrate the superiority of our model in extracting distinct patterns across neurodegenerative disorder stages. Furthermore, these two modules are compatible and complementary. By integrating both modules, our smBGC achieves the best performance.

\begin{table}[ht]
	\centering
	\caption{Results of smBGC variants w/o each proposed component on the PPMI dataset.}
	\label{ablation}
 \setlength{\tabcolsep}{3pt} % Adjust column separation	
 \ssmall
  \resizebox{0.5\textwidth}{!}{
\begin{tabular}{ccccccc}
\toprule
\multicolumn{2}{c}{\multirow{2}*{Methods}}& \multicolumn{5}{c}{NC vs. Prodromal vs. PD}  \\\cmidrule{3-7}
& & ACC & F1 & AUC & SEN & SPE  \\
\midrule
& w/o CFD \& ASFE         & 41.73 $\pm$ 5.26 & 36.97 $\pm$ 5.87 & 56.50 $\pm$ 4.08 & 39.53 $\pm$ 3.85 & 69.09 $\pm$ 5.43         \\
& w/o ASFE        & 57.44 $\pm$ 5.29 & 50.15 $\pm$ 1.59 & 70.94 $\pm$ 6.76 & 52.24 $\pm$ 5.63  & 80.88 $\pm$ 2.94       \\
& w/o CFD         & \underline{83.24 $\pm$ 2.29} & \underline{78.28 $\pm$ 3.61} & \underline{97.47 $\pm$ 1.19} & \underline{82.96 $\pm$ 2.64} & \underline{91.73 $\pm$ 5.91}      \\ \midrule
& \textbf{smBGC (ours)} & \textbf{91.15 $\pm$ 3.19 } & \textbf{91.24 $\pm$ 3.13} & \textbf{96.27 $\pm$ 0.91} & \textbf{91.77 $\pm$ 4.33} & \textbf{95.61 $\pm$ 3.52}  \\ \bottomrule
\end{tabular}}
\end{table}

\subsubsection{Analysis of Different Modalities}

We evaluate the model performance when different modalities are utilized for enhancement. Specifically, in both steps of the stage-aware feature enhancement strategy, it is necessary to compute the feature scoring matrix: the common scoring matrix $S_c$ for the first step and the stage scoring matrix $F_k$ for the second step. Since fMRI and DTI data are collected for each subject, the scoring matrix $S$ can be computed for each modality, resulting in $F_k^f$ for the functional modality and $F_k^s$ for the structural modality. Four possible combinations for the two steps are: $f$+$f$, $f$+$s$, $s$+$s$, and $s$+$f$ (where $f$ represents the functional modality and $s$ represents the structural modality). Fig. \ref{fig:modal} illustrates the accuracy results for classification tasks under different combinations. It can be observed that all combinations can improve model performance. However, the performance of single-modality combinations is consistently lower than that of multimodal combinations. This demonstrates that the complementary information between modalities plays a significant role in the enhancement strategy. 

 \begin{figure}[!t]
	\centering
	\includegraphics[width=\linewidth]{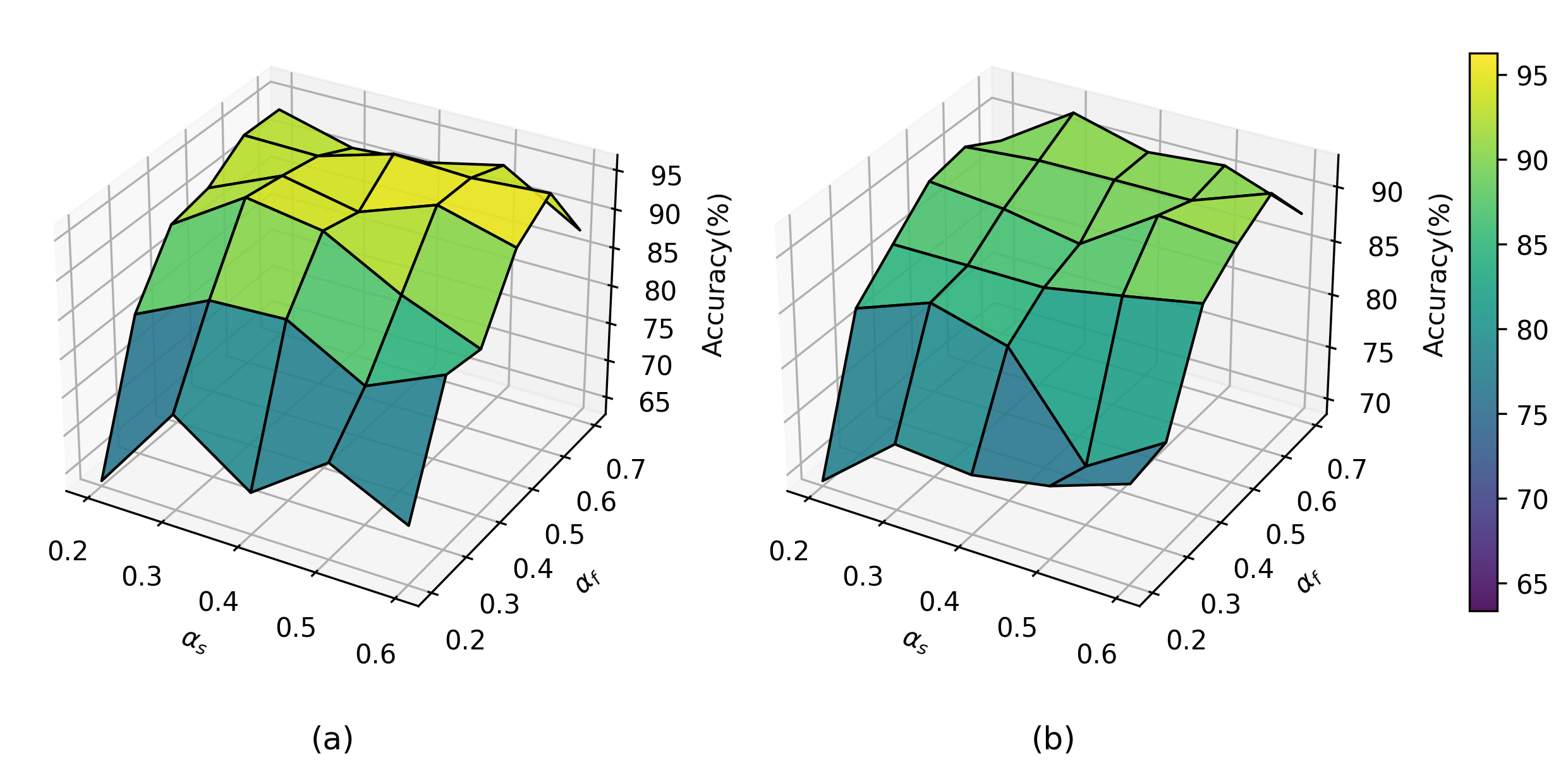}
	\caption{The classification accuracy of smBGC with varied different hyperparameters on (a) ADNI-3 with three stages and (b) PPMI datasets.}
	\label{para}
\end{figure}

\begin{figure*}[h!]
\centering
\subfloat{
 \begin{minipage}[t]{0.32\textwidth}
	\includegraphics[width=6cm]{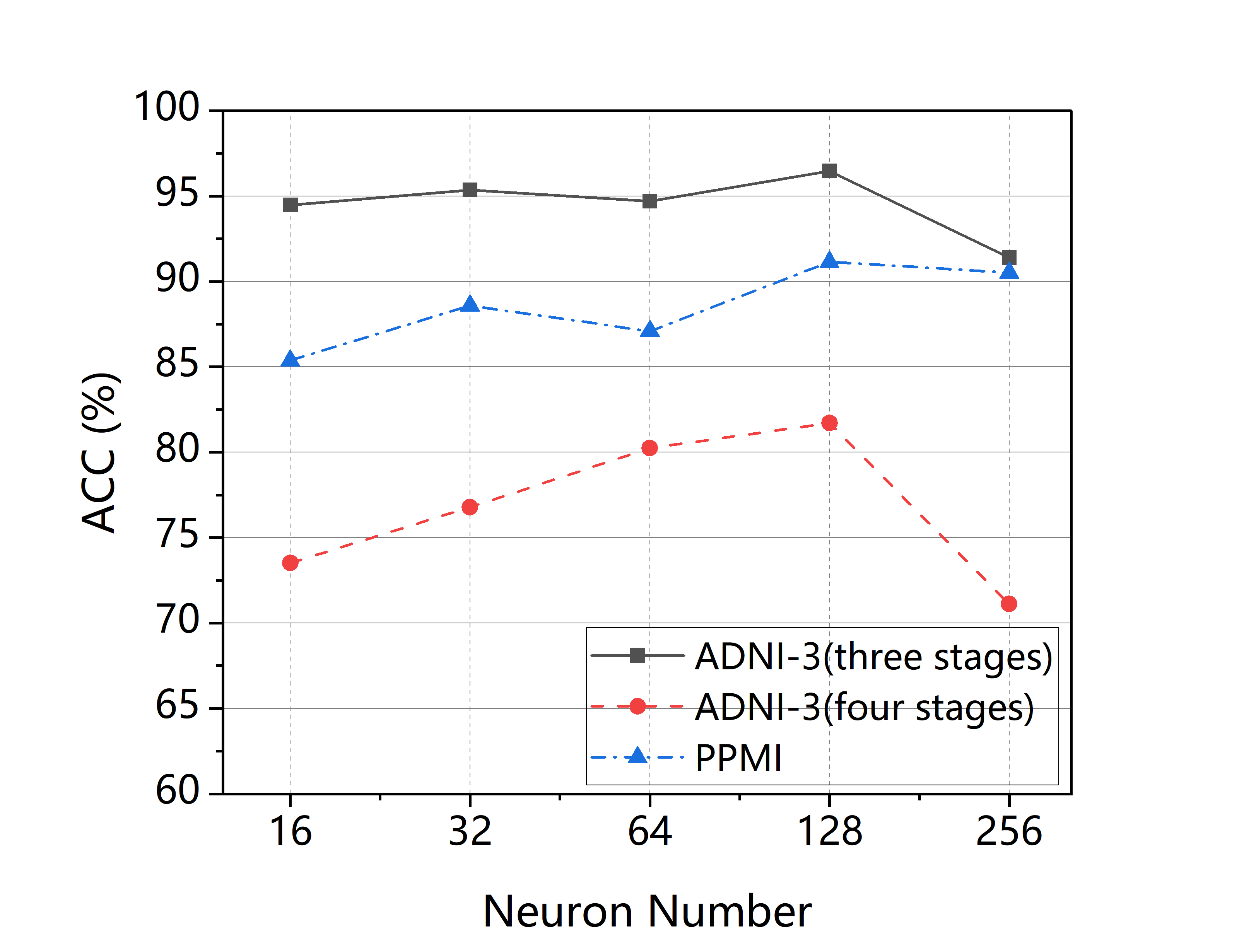}% This is a *.eps file
\end{minipage}}
\subfloat{
 \begin{minipage}[t]{0.32\textwidth}
	\includegraphics[width=6cm]{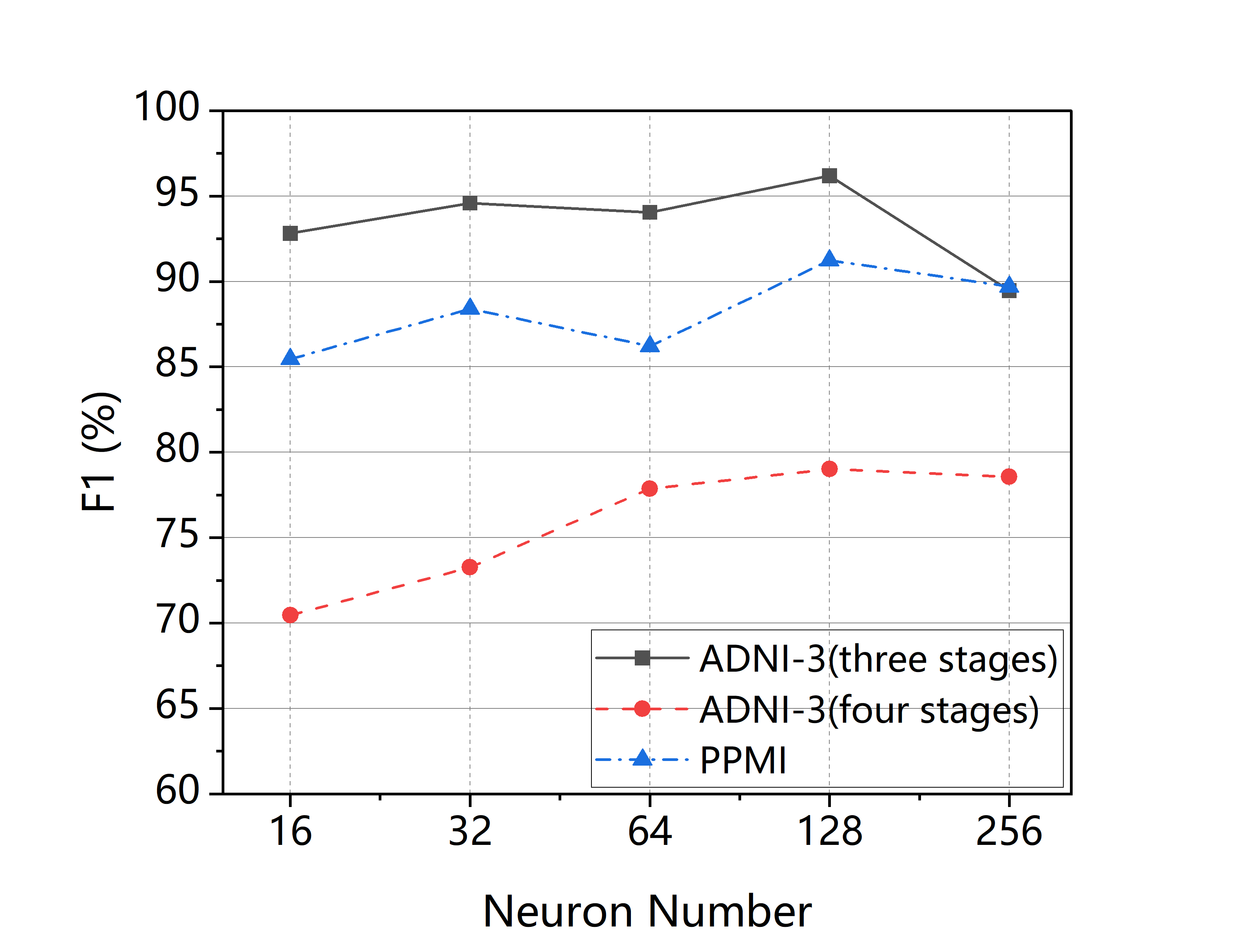}% This is a *.eps file
\end{minipage}}
\subfloat{
 \begin{minipage}[t]{0.32\textwidth}
	\includegraphics[width=6cm]{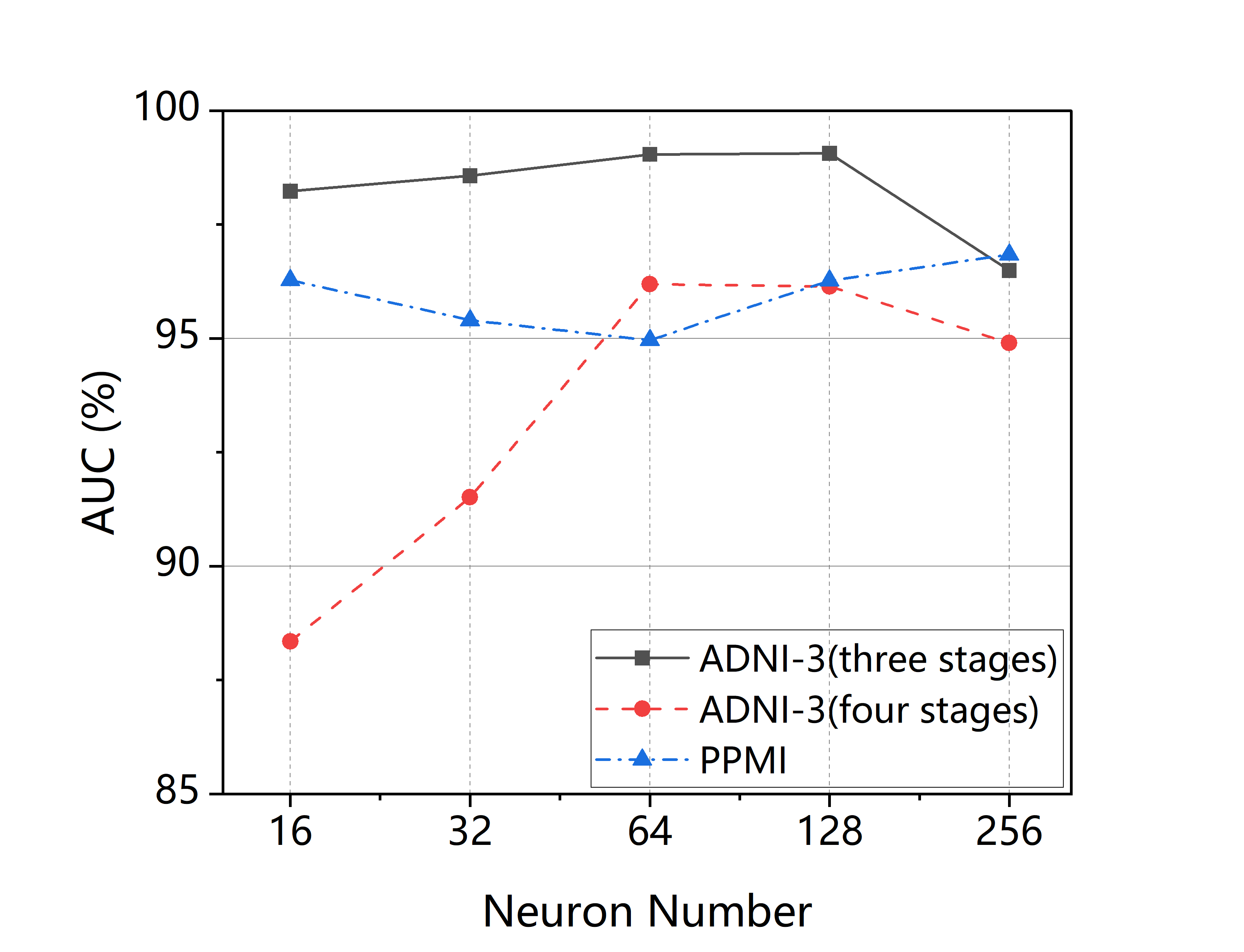}% This is a *.eps file
\end{minipage}}
	\caption{Accuracy, F1-score, and AUC values of the proposed model under different neuron numbers over three datasets.}
	\label{fig:neuron}
\end{figure*}

\subsection{Parameters Analysis}

\subsubsection{The Magnitude of Feature Enhancement}

We perform a parameter analysis of the key hyperparameters in smBGC, specifically the hyperparameters $\alpha_f$ and $\alpha_s$, which determine the augmentation strength for structural and functional views, respectively, during graph augmentation. By varying these parameters from 0.2 to 0.7, we evaluate the model performance on the three-stage classification task using ADNI-3 and PPMI datasets. The results are shown in Fig. \ref{para}. It can be observed that the classification accuracy remains relatively stable when the parameter values are changed. This indicates that, overall, our model is robust to parameter perturbations, demonstrating its insensitivity to these hyperparameters. However, when the parameter values are set too small, noise features within the brain graphs remain significant, and the stage-specific characteristics of brain disorders are not fully revealed. In such cases, the model struggles to extract meaningful information from the graph representations.

\subsubsection{Neuron Number in Hidden Layer}

We investigate the impact of the number of neurons in the hidden layer on the performance of smBGC. Specifically, we conduct experiments with varying hidden layer sizes. As is shown in Fig. \ref{fig:neuron}, the results reveal a clear trend: as the number of neurons in the hidden layer increases, the overall performance across the three tasks improves. Notably, smBGC achieves its best performance when the hidden layer is configured with 128 neurons. This observation highlights two key aspects. First, it underscores the model's stability under different neuron configurations, demonstrating its robustness in handling varying setups. Second, it suggests that the choice of hidden layer size plays an important role in maximizing the model's potential for learning meaningful representations.

\section{Conclusion}\label{conclusion}

This paper has presented a novel Stage-aware Multimodal Brain Graph Contrastive Learning (smBGC) method, which effectively distinguishes different stages of neurodegenerative disorders by exploring distinct
brain patterns within each stage. In particular, the proposed method utilizes a stage-aware feature enhancement strategy, removing noise features and distilling stage-specific features to achieve precise targeting of different stages of disorder. smBGC explores the stage-specific characteristics of the progression of neurodegenerative disorders and provides deep insight into pathological mechanisms. It shows great potential to predict the progression trajectory of neurodegenerative disorders, which is crucial for early intervention and treatment planning. Extensive experimental results on two real-world datasets, i.e., ADNI-3 and PPMI, have demonstrated that smBGC outperforms state-of-the-art methods on Alzheimer’s disease detection and Parkinson’s disease detection.

\bibliographystyle{IEEEtran}
\bibliography{ref.bib}

\end{document}